\documentclass{article} 
\usepackage[final]{colm2026_conference}

\usepackage{microtype}
\usepackage{hyperref}
\usepackage{url}
\usepackage{booktabs}
\usepackage{graphicx}
\usepackage[table]{xcolor}
\usepackage{makecell}
\usepackage{subcaption}
\usepackage{nicematrix}
\usepackage[most]{tcolorbox}
\usepackage{fvextra}
\usepackage[T2A,T1]{fontenc}  
\usepackage[russian,english]{babel}  
\usepackage{tabularx}
\usepackage{booktabs}  
\usepackage{bbding}

\usepackage{lineno}

\definecolor{darkblue}{rgb}{0, 0, 0.5}
\definecolor{beige}{RGB}{245, 235, 220}
\definecolor{lightgray2}{RGB}{220, 220, 220}
\hypersetup{colorlinks=true, citecolor=darkblue, linkcolor=darkblue, urlcolor=darkblue}

\newcommand{\ourbench}{\textsc{CresOWLve}}
\newcommand\owl{\raisebox{-8pt}{\includegraphics[width=2em]{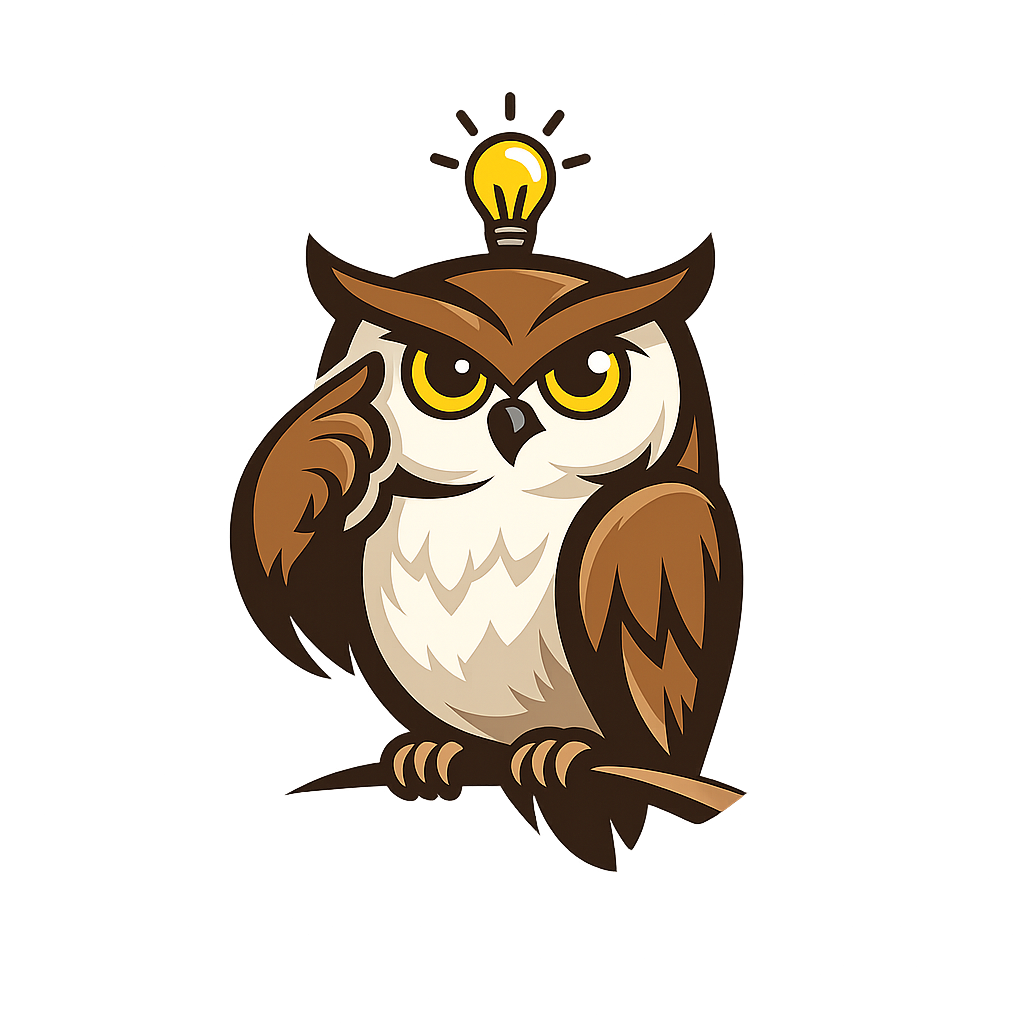}}}

\newtcolorbox[list inside=prompt,auto counter,number within=section]{prompt}[1][]{
    colbacktitle=black!60,
    fonttitle=\small,
    coltitle=white,
    fontupper=\footnotesize,
    boxsep=3pt,
    left=0pt,
    right=0pt,
    top=0pt,
    bottom=0pt,
    boxrule=1pt,
    #1,
    breakable,              
}

\usepackage{tcolorbox}
\usepackage{xcolor}
\usepackage{enumitem}
\usepackage{listings}
\usepackage{inconsolata}

\definecolor{lightgraybg}{HTML}{F7F7F7}
\definecolor{lightbluebg}{HTML}{EEF5FF}
\definecolor{lightgreenbg}{HTML}{EEF9F0}
\definecolor{lightredbg}{HTML}{FFF1F1}
\definecolor{bordergray}{HTML}{D9D9D9}
\definecolor{darkblue}{HTML}{1F4E79}
\definecolor{darkgreen}{HTML}{2E7D32}
\definecolor{darkred}{HTML}{B23A48}

\lstdefinestyle{appendixjson}{
  basicstyle=\ttfamily\small,
  breaklines=true,
  columns=fullflexible,
  keepspaces=true,
  frame=single,
  framerule=0.3pt,
  rulecolor=\color{bordergray},
  backgroundcolor=\color{lightgraybg},
  showstringspaces=false,
  xleftmargin=6pt,
  xrightmargin=6pt,
  aboveskip=6pt,
  belowskip=6pt
}

\title{\owl{} \ourbench{}: Benchmarking Creative Problem-Solving Over Real-World Knowledge}

\newcommand{\authsep}{\hspace{1.2em}}
\newcommand{\camready}[1]{{#1}}

\author{Mete Ismayilzada\textsuperscript{1,2} \authsep Renqing Cuomao\textsuperscript{*1} \authsep Daniil Yurshevich\textsuperscript{*1} \authsep Anna Sotnikova\textsuperscript{1} \\ 
\textbf{Lonneke van der Plas}\textsuperscript{\textdagger2} \authsep \textbf{Antoine Bosselut}\textsuperscript{\textdagger1} \\
\textsuperscript{1}EPFL, \textsuperscript{2}Università della Svizzera italiana (USI)
}

\begingroup
\footnotetext{Equal contribution}
\endgroup

\begingroup
\footnotetext{Equal supervision}
\endgroup

\begin{document}

\ifcolmsubmission
\linenumbers
\fi

\maketitle

\begin{abstract}
Creative problem-solving requires combining multiple cognitive abilities, including logical reasoning, lateral thinking, analogy-making, and commonsense knowledge, to discover insights that connect seemingly unrelated pieces of information. However, most existing benchmarks for large language models (LLMs) evaluate only specific components of this process. Moreover, many creativity-oriented benchmarks rely on artificially constructed brainteasers or contrived scenarios that do not reflect how creative problem-solving occurs in real-world settings. To address this gap, we introduce \ourbench{} (containing 2,061 examples across five difficulty levels), a benchmark for evaluating creative problem-solving using puzzles grounded in real-world knowledge. Problems in \ourbench{} require employing multiple creative thinking strategies, retrieving facts from diverse domains, and creatively combining them to arrive at a solution. Evaluating several frontier non-thinking and thinking LLMs, we show that \ourbench{} remains highly challenging. Our analysis reveals a consistent performance gap: models perform substantially better on factual questions than on creative ones (up to $-17\%$ drop). While models can often retrieve the relevant knowledge, they struggle to form the non-obvious creative connections required to integrate the knowledge and arrive at the correct answer.

\end{abstract}

\section{Introduction}
Creative problem-solving is a central component of human intelligence and plays a crucial role in scientific discovery, innovation, and everyday reasoning \citep{duncker1948problem}. Solving problems creatively often requires a combination of several cognitive processes \citep{ismayilzada2024creativity}. \textit{Vertical thinking} refers to systematic, step-by-step reasoning that derives conclusions through logical progression, while \textit{lateral thinking} involves restructuring a problem and identifying non-obvious connections between seemingly unrelated ideas \citep{de1970lateral, waks1997lateral}. Similarly, creativity research distinguishes between \textit{divergent thinking}, which emphasizes generating multiple possible ideas or solutions, and \textit{convergent thinking}, which focuses on identifying a single insightful solution by integrating different pieces of information \citep{guilford1967nature}. Creative problem solving also relies on additional abilities such as abstraction and analogy making \citep{hofstadter2001analogy}, which allow individuals to transfer knowledge across domains, as well as commonsense reasoning \citep{davis2015commonsense}, which provides background knowledge about how the world works. \

\begin{figure}[t]
\includegraphics[width=\linewidth,scale=0.5]{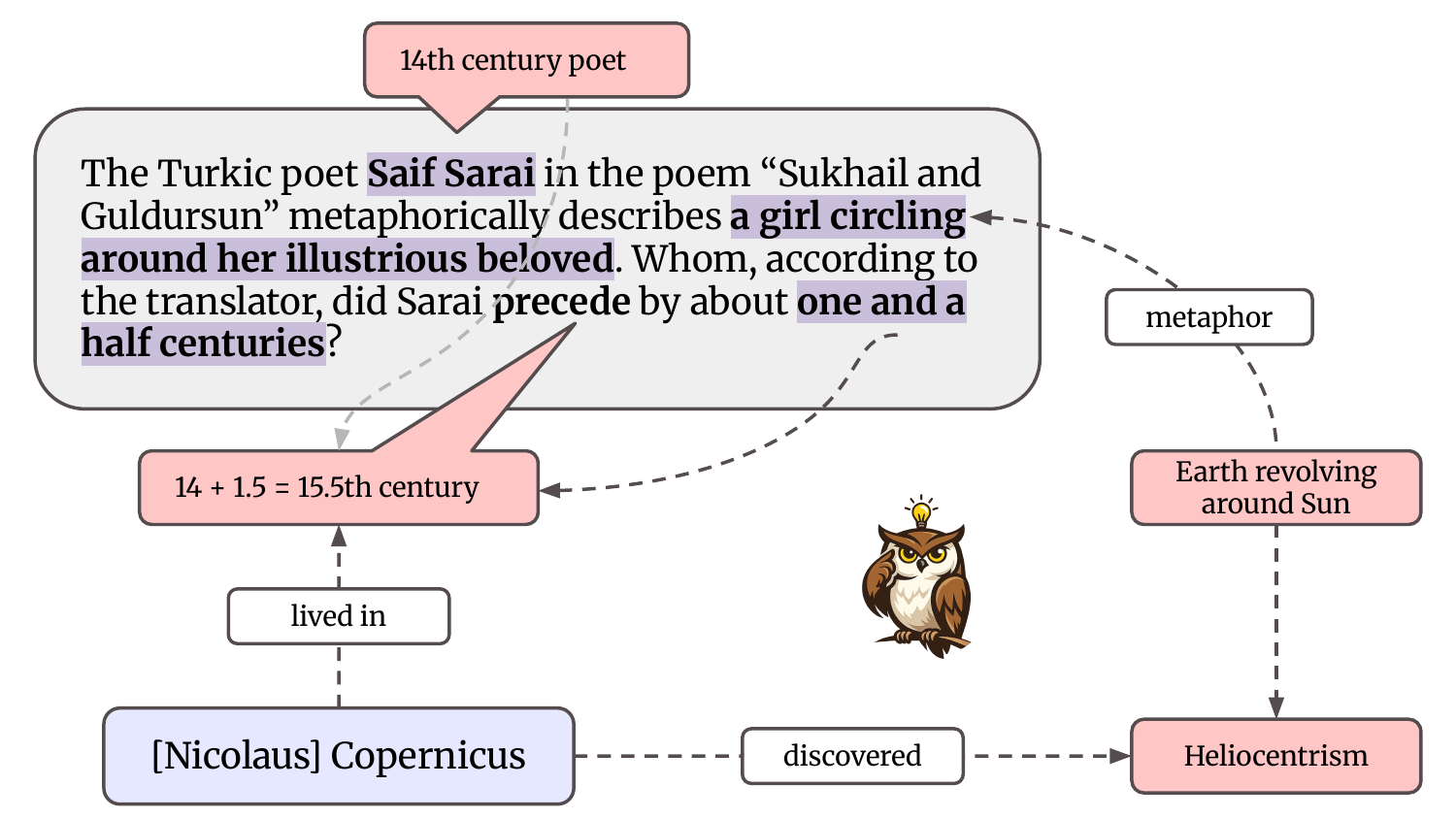}
\centering
\caption{An example from \ourbench{} annotated with the real-world knowledge and creative thinking strategy.}
\label{fig:main-figure}
\end{figure}

Despite rapid advances in large language models (LLMs) \citep{zhao2023survey}, most existing benchmarks evaluate only specific aspects of creative problem solving. Some benchmarks emphasize vertical reasoning through tasks such as factual reasoning \citep{hendrycks2020measuring, romanou2024include}, mathematical/logical inference \citep{hendrycks2021measuring, srivastava2022beyond}, and commonsense reasoning \citep{sakaguchi2021winogrande, talmor2019commonsenseqa, lin2021riddlesense}, while others focus on lateral thinking using brainteasers and situational puzzles \citep{han2025creativity, chen2024weak, huang2024lateval, jiang2023brainteaser}. 
Other works have proposed benchmarks to measure convergent/divergent thinking using psychometric tests \citep{stevenson2022putting, goes2023pushing, bellemare2024divergent} or puzzles and games \citep{tian2024macgyver, alavi2023large}. 
Although these benchmarks capture important individual capabilities, real-world creative problem-solving typically requires a combination of these skills. Furthermore, many existing evaluations rely on artificially constructed brainteasers (e.g. \textit{``What type of cheese is made backwards?''}) or contrived situational scenarios (e.g. \textit{``Two men are beating each other up and both of them suddenly fall to the floor. How?''}). In practice, however, creative insights often emerge from retrieving relevant knowledge across diverse domains and reasoning about it in novel ways. As a result, these benchmarks provide only a partial picture of the real-world creative problem-solving abilities of models.

To address this gap, we introduce \ourbench{}, a benchmark designed to evaluate creative problem-solving in LLMs through single-answer puzzles that require the combination of multiple cognitive abilities, such as vertical/lateral, convergent/divergent thinking, analogy-making, and are grounded in real-world knowledge (Appendix Table \ref{tab:cps-benchmarks} provides a full comparison to past relevant benchmarks and Figure \ref{fig:main-figure} illustrates an example puzzle from our benchmark. See Appendix Table \ref{tab:benchmark-examples} for more examples with varying difficulty levels). \ourbench{} is constructed from questions drawn from the renowned Russian intellectual game ``What? Where? When?'' \footnote{en.wikipedia.org/wiki/What\%3F\_Where\%3F\_When\%3F}, in which expert human participants solve carefully crafted problems that require not only broad world knowledge but also creative insight to combine disparate facts in non-obvious ways. 
\camready{Originally televised in the mid-1970s and still running today, the game features teams of six expert players who have 60 seconds to deliberate and produce a single answer to questions authored by viewers and difficulty-rated by an editorial board, with thousands of validated items archived in the public database that we draw from.}
To ensure accessibility and relevance, we design a multi-stage benchmark construction pipeline that filters unsuitable and non-creative questions and translates the remaining puzzles into English with manual validation. The resulting dataset provides a diverse and high-quality benchmark for evaluating creative problem-solving grounded in real-world knowledge
\footnote{https://huggingface.co/datasets/mismayil/cresowlve}. 
We summarize our main contributions as follows: 
\textbf{1)} We present \ourbench{}, a bilingual benchmark for creative problem-solving grounded in real-world knowledge and solvable by human experts. \ourbench{} spans a diverse range of knowledge and creative domains, varies in difficulty, requires multiple creative thinking strategies and is manually validated to ensure quality. \textbf{2)} We evaluate several frontier open-weight and proprietary LLMs varying in their size and reasoning mode, on \ourbench{} and demonstrate that, despite recent advances, the benchmark remains highly challenging, revealing substantial gaps in model performance on creative reasoning over real-world knowledge. Notably, thinking models substantially outperform their non-thinking counterparts, demonstrating that extended reasoning is a key enabler for creative problem-solving. \textbf{3)} We conduct a comprehensive analysis of model performance across multiple dimensions, including question difficulty, domains of knowledge and creativity, and show that models consistently underperform on creative questions relative to factual ones. Our error analysis also shows that while models can often retrieve the relevant knowledge, they frequently fail to make the creative connections among facts necessary to arrive at the correct answer.

\label{sec:introduction}

\section{Related Work}

\paragraph{Creativity Evaluation}
Evaluating creativity is inherently challenging due to its subjective nature, and traditional assessments often rely on human judgment using techniques such as the Consensual Assessment Technique \citep{ismayilzada2024creativity, amabile1983social} or psychometric tests \citep{stevenson2022putting, goes2023pushing, guilford1967nature, mednick1962associative}. To complement these approaches, several computational metrics and corresponding benchmarks have been proposed that capture different dimensions of creativity, including novelty \citep{zhang2025noveltybench, lu2024ai, organisciak2023beyond, johnson2023divergent}, diversity \citep{padmakumar2023does, jiang2026artificial}, surprise \citep{bunescu-uduehi-2022-distribution, karampiperis2014towards, itti2009bayesian}, and quality \citep{franceschelli2025thinking, franceschelli2022deepcreativity}. Other works have introduced multi-task creativity evaluation datasets \citep{ismayilzada2025creative, hou2025creativityprism, xue2025crebench} where problems don't have a single correct answer, but rather many subjective answers, and the evaluation focuses on measuring different dimensions of output creativity. In creative problem-solving, however, there are often single (or a few) correct answers, and the emphasis is on the type of creative thinking process involved to get to the final answer. Our work similarly targets creative problem solving with a single correct answer, hence our analysis focuses more on the evaluation of different creative thinking abilities employed by the model to solve our benchmark puzzles.

\paragraph{Creative Problem-Solving Benchmarks}
While most works have focused on measuring vertical thinking capabilities of LLMs through tasks such as factual reasoning \citep{hendrycks2020measuring, romanou2024include}, mathematical/logical inference \citep{hendrycks2021measuring, srivastava2022beyond}, and commonsense reasoning \citep{sakaguchi2021winogrande, talmor2019commonsenseqa, bisk2020piqa, ismayilzada2023crow, lin2021riddlesense}, several benchmarks have also been proposed to evaluate other aspects of problem-solving that requires more creativity \citep{ismayilzada2024creativity}. More specifically, past work has evaluated lateral thinking using mostly brainteasers and situational puzzles \citep{han2025creativity, chen2024weak, huang2024lateval, jiang2023brainteaser, kraaijveld2025columbus, todd2024missed}, convergent/divergent thinking using psychometric tests \citep{stevenson2022putting, goes2023pushing, bellemare2024divergent} or puzzles and games \citep{tian2024macgyver, alavi2023large, wadhwa2026createtestingllmsassociative} and abstraction/analogy-making through synthetic puzzles \citep{ahrabian2024curious, chollet2019measure, moskvichev2023conceptarc, lewis2024using}. Other works have focused on specific domains such as coding \citep{lu2025benchmarking}, and mathematics \citep{ye2025assessing, sun2025omega}. Our benchmark on the other hand requires models to employ multiple creative thinking abilities and reason creatively over real-world knowledge drawn from diverse domains. We also note that while the questions from the ``What?Where?When'' game has been used for LLM evaluation in the past by \citet{lifar2024llama}, our work significantly differs in several aspects: 1) \citet{lifar2024llama} only considers questions with more factuality and shorter reasoning chains while we focus on more creative questions 2) \citet{lifar2024llama} does not perform any data filtering to remove unanswerable or Russian-culture specific questions 3) \citet{lifar2024llama} evaluates only one LLM (namely, \texttt{LlaMa3-405B}) and provides limited analysis while we benchmark several frontier LLMs and provide extensive analysis on question difficulty, reasoning types and error categories. 4) \citet{lifar2024llama} considers Russian-only evaluation, while we prepare and evaluate models on a human-validated English version of the dataset as well.
\label{sec:related-work}

\section{\ourbench}
\subsection{Benchmark Construction}
\paragraph{Data Collection}
\label{sec:data_collection}
We collect $3,789$ questions from the public database at \texttt{db.chgk.info}, which contains questions from the popular Russian intellectual game ``What?Where?When?'' spanning over 50 years. Questions in this game have been manually crafted by humans and often require a combination of skills such as logical thinking, intuition, and creative insight. Each question is annotated with a short answer, an answer explanation, and a difficulty rating (1 to 5) manually assigned by the game organizers. 
To ensure diversity in benchmark difficulty, 
we collect $\approx 700$ questions per difficulty rating. We perform several filtering and annotation stages using \texttt{GPT-4o} \citep{hurst2024gpt} due to its strong performance \citep{gpt4annot,gilardi2023chatgpt}. 
Prompts can be found in Appendix Tables \ref{tab:prompt-ext-mat}, \ref{tab:prompt-ru-spec}, \ref{tab:prompt-translation}, \ref{tab:prompt-reasoning-type}, \ref{tab:prompt-domain} and \ref{tab:prompt-culture-lang}. Example questions for each stage can be found in Appendix Table \ref{tab:pipeline-examples}. 
\camready{We note that \texttt{GPT-4o} is used solely for data annotation and filtering during benchmark construction, and is not included among the evaluated models, avoiding any potential self-preference bias in our results. To assess the reliability of these annotations, we additionally conducted an inter-annotator agreement study on a sample of items independently re-annotated by two of the authors, finding moderate-to-strong agreement across all annotation fields (Appendix \ref{sec:annot-reliability}).}

\paragraph{Data Filtering}
\label{sec:data_filtering}
Since the original game is in Russian and also in-person interactive, some questions rely heavily on linguistic and cultural knowledge specific to Russia, and some require physical inspection of external material, such as images or handouts. Hence, these questions are either extremely hard to answer or unanswerable completely. Therefore, we perform several filtering steps to ensure the validity and wide accessibility of the questions. 
We first filter out unanswerable questions by annotating each question on whether they require physical external materials to be solved. This step removes $295$ samples.
Next, we filter out questions that depend heavily on knowing specific facts rooted deeply in the Russian language and culture, rendering them extremely hard to answer or untranslatable into English. This step removes $799$ samples.

\paragraph{Translation}
\label{sec:translation}
Our goal in this work is to design a benchmark that is relevant and accessible for measuring creative problem-solving in frontier LLMs. However, since most LLMs are more proficient reasoners in English than in other languages, we also prepare an English version of our benchmark by translating each question-answer pair with all its metadata.

\begin{figure}[h]
    \centering
    \begin{minipage}[t]{0.55\textwidth}
        \vspace{0pt}
        \centering
        \includegraphics[width=\linewidth, trim={1cm 1cm 0cm 0cm},clip]{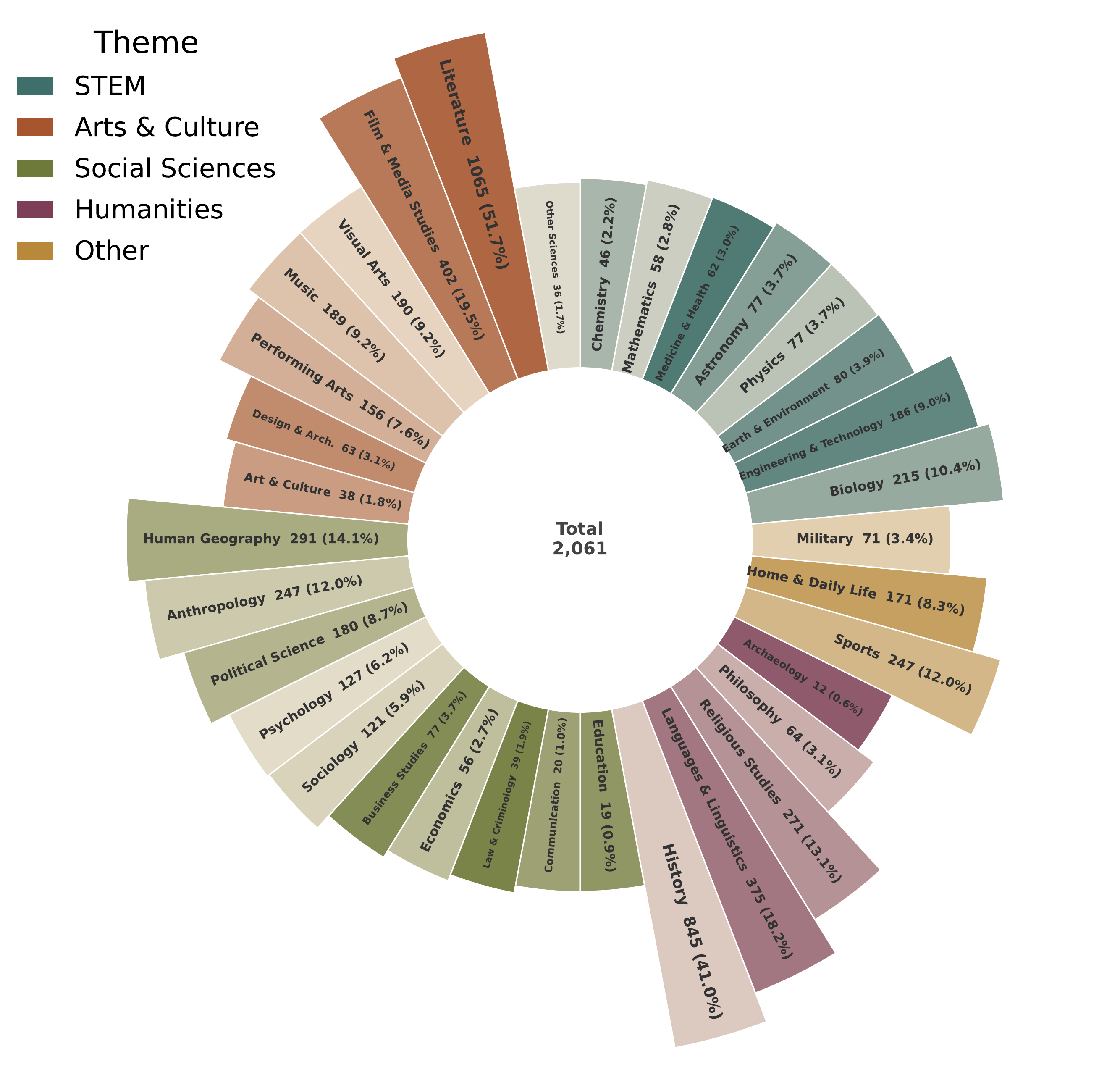}
        \subcaption{Distribution of knowledge domains.\label{fig:domains-dist}}
    \end{minipage}
    \hfill
    \begin{minipage}[t]{0.44\textwidth}
        \vspace{0pt}
        \centering
        \includegraphics[width=\linewidth]{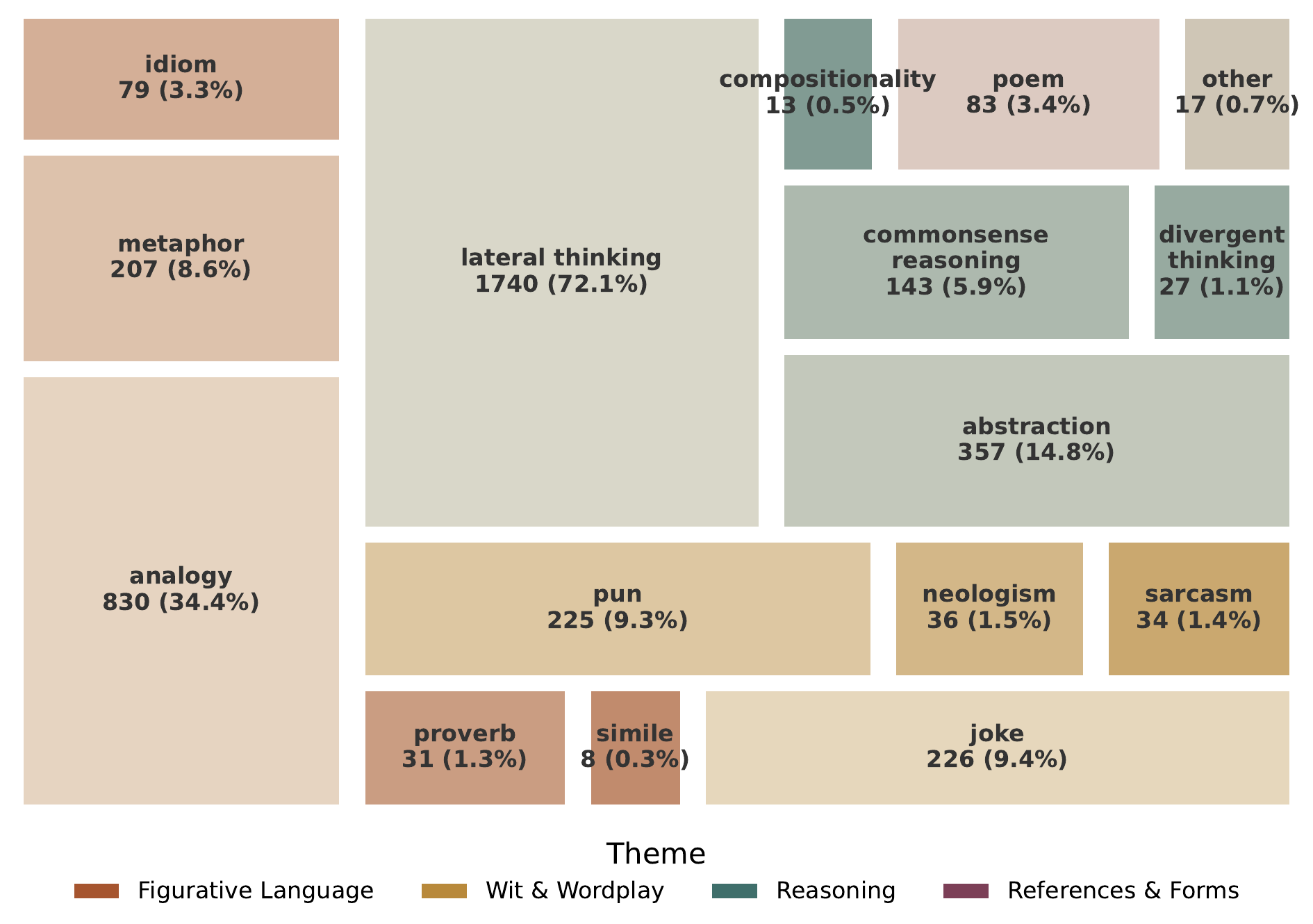}
        \subcaption{Distribution of creative domains.\label{fig:creativity-dist}}
        \vspace{0.5em}
        \includegraphics[width=\linewidth]{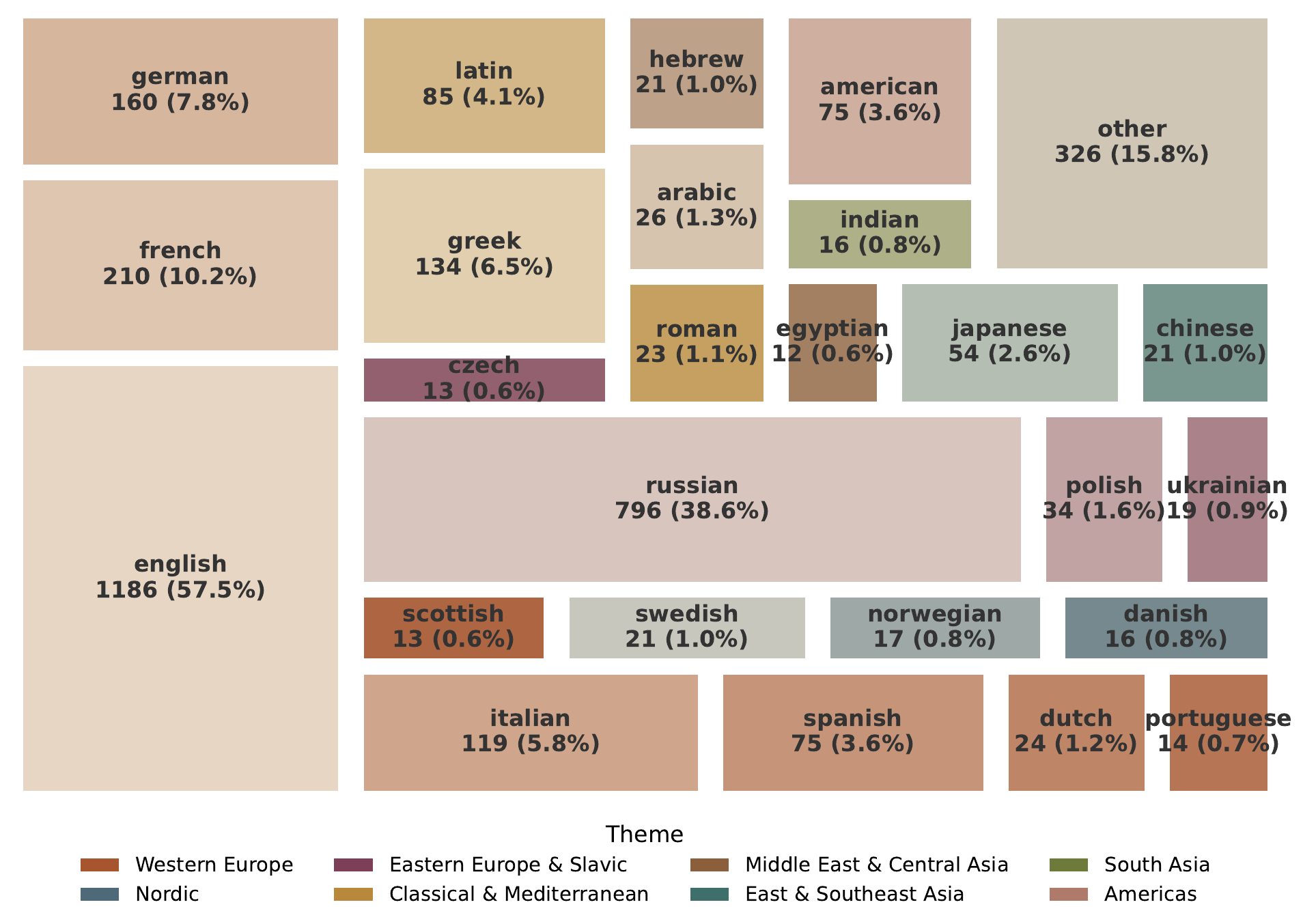}
\        \subcaption{Distribution of cultures/demographics.\label{fig:langs-dist}}
    \end{minipage}
    \caption{Diversity of real-world knowledge, creative language, and cultures in \ourbench{}.}
    \label{fig:domains-creative-cultures}
\end{figure}

\paragraph{Human Validation}
\label{sec:human_validation}
While LLMs have recently become remarkably effective in data annotation tasks, they can still exhibit hallucination and reasoning failures \citep{tan2024large}. Therefore, we conduct a final human validation on the entire benchmark to ensure high quality. More specifically, three authors of this paper reviewed all remaining questions from the last filtering step to validate that they are answerable, not Russian-specific, and that the translations are correct. This step further removed $282$ samples from the benchmark, leaving $2,413$ samples for final evaluation.

\paragraph{Creative vs. Factual Reasoning}
\label{sec:creative_vs_factual}
As noted by \citet{lifar2024llama}, some questions can be answered with purely factual retrieval and reasoning involving no to minimal creative leap-of-thought. To further distinguish this dimension, we automatically categorize questions as either \textit{factual} or \textit{creative}, where creative questions require the model to combine and apply knowledge in ways that go beyond direct retrieval. This annotation yields  $352$ factual and $2,061$ creative samples. We later use this factual subset to analyze the performance on creative vs. factual reasoning questions.

\subsection{Final Benchmark}
\label{sec:final_benchmark}
Our final benchmark contains $2,061$ samples (creative subset) both in Russian and English (referred to as \ourbench{}-Ru and \ourbench{}-En respectively). Data statistics about the benchmark can be found in Appendix Table \ref{tab:data-stats}. Due to the subjective nature of creativity evaluation, designing such benchmarks at scale is challenging; we therefore source our puzzles from human-authored material, though we additionally propose and prototype a pipeline for synthetic \ourbench{}-style construction from real-world knowledge graphs (Appendix~\ref{sec:synthetic}).

\paragraph{Knowledge Domains}
\label{sec:knowledge_domains}
As noted earlier, one of the challenges of this benchmark is to creatively reason over real-world knowledge across domains. To visualize the domain diversity, we automatically annotate each benchmark sample with the knowledge domains required to solve it. The initial annotation yields $541$ domains, which are too fine-grained and contain substantial overlap. We then consolidate them into broader subject domains, resulting in $34$ coarse-grained, largely non-overlapping categories. Figure \ref{fig:domains-dist} illustrates the coarse domain breakdown. We can see that while the benchmark covers a wide range of topics, Literature, History, Film \& Media Studies, and Languages \& Linguistics are the dominating subjects — consistent with the source material, as creative problem-solving in human intellectual games concentrates naturally in domains where figurative language, narrative, and cultural reference offer rich material for non-obvious connections. We also note that each question is annotated with multiple topics, and most questions involve at least two to four topics (See Appendix Figure \ref{fig:num-domains}).

\paragraph{Creative Language \& Thinking}
\label{sec:creative_domains}
In addition to the knowledge domains, our benchmark questions also require reasoning about creative language and employing multiple creative thinking strategies such as lateral thinking, abstraction, and analogy-making. To quantify them, we perform an automatic annotation and report the breakdown in Figure \ref{fig:creativity-dist}. We see that most of the questions require lateral thinking, an ability to identify non-obvious associations, and a substantial number of questions involve abstraction and analogy-making. Many questions also require reasoning about jokes, puns, or metaphors, as well as commonsense knowledge, and most questions involve two creative domains (Appendix Figure \ref{fig:num-concepts}).

\paragraph{Cultures \& Demographics}
\label{sec:cultures}
While the original game is in the Russian language, its questions often involve knowledge about entities and people from other cultures, too. To quantify the diversity of cultures and demographics in our benchmark, we automatically annotate each sample in the benchmark with the cultures involved in solving the question. We report the breakdown in Figure \ref{fig:langs-dist}. We note that while English and Russian cultures dominate, our benchmark also contains a substantial number of questions on other cultures, such as French, German, Italian, and Greek. Similar to knowledge and creative domains, most questions often involve knowledge about more than one culture (Appendix Figure \ref{fig:num-langs}).
\label{sec:benchmark}

\section{Experimental Setup}
\paragraph{Models}
\label{sec:models}
We assess a broad set of LLMs on our benchmark, varying in reasoning mode, architecture, size, and training paradigm. Since our benchmark requires complex reasoning, we particularly distinguish between \textit{non-thinking} models, which generate responses directly without explicit intermediate reasoning (unless instructed to via Chain-of-Thought prompting \citep{wei2022chain}) and \textit{thinking} models, which are trained to explicitly think with a certain amount of effort before producing a final answer. We consider the following non-thinking models: \texttt{GPT-4.1-mini}, \texttt{GPT-4.1} \citep{achiam2023gpt}, \texttt{OLMo-2-32B-Instruct} \citep{olmo20242}, \texttt{Qwen3-235B-A22B-Instruct} \citep{yang2025qwen3}, \texttt{C4AI-Command-A} \citep{cohere2025command}, \texttt{Mistral-Large-3-675B-Instruct} \citep{mistral3large} and \texttt{Llama-3.3-70B-Instruct} \citep{grattafiori2024llama} and following thinking models: \texttt{Qwen3.5-397B-A17B} \citep{qwen3.5}, \texttt{Qwen3-235B-A22B-Thinking} \citep{yang2025qwen3}, \texttt{DeepSeek-V3.2} \citep{liu2025deepseek}, \texttt{GLM-5} \citep{zeng2026glm}, \texttt{GPT-5.4} \citep{gpt54openai}, \texttt{Gemini-3-Flash} \citep{gemini3google}, and \texttt{Gemini-3.1-Pro} \citep{gemini31google}.

\rowcolors{1}{}{}
\begin{table}[t]
\centering
\small
\scalebox{0.8}{
\begin{NiceTabular}{lccccc}
\toprule
 & & \multicolumn{2}{c}{\ourbench{}-En} & \multicolumn{2}{c}{\ourbench{}-Ru} \\
\cmidrule(lr){3-4} \cmidrule(lr){5-6}
\textbf{Model} & \textbf{Thinking Effort} & \textbf{Exact Match} & \textbf{LLM Judge} & \textbf{Exact Match} & \textbf{LLM Judge} \\
\midrule
\Block[fill=beige]{1-6}{\textit{Non-Thinking Models}} \\
\midrule
\Block[fill=beige]{7-6}{}
\textsc{OLMo-2-32B-Instruct} & - & 2.09 & 5.29 & 0.44 & 1.31 \\
\textsc{C4AI-command-a} & - & 4.46 & 9.07 & 3.74 & 7.08 \\
\textsc{Llama-3.3-70B-Instruct} & - & 5.05 & 10.14 & 3.64 & 7.04 \\
\textsc{Mistral-Large-3-675B-Instruct} & - & 11.26 & 17.61 & 15.48 & 22.22 \\
\textsc{Qwen3-235B-A22B-Instruct} & - & 13.05 & 20.82 & 13.54 & 20.14 \\
\textsc{GPT-4.1-mini} & - & 7.33 & 13.78 & 4.85 & 8.83 \\
\textsc{GPT-4.1} & - & \textbf{13.44} & \textbf{24.99} & \textbf{17.03} & \textbf{26.06} \\
\midrule
\Block[fill=lightgray2]{1-6}{\textit{Thinking Models}} \\
\midrule
\Block[fill=lightgray2]{11-6}{}
\textsc{Qwen3.5-397B-A17B} & adaptive & 19.36 & 31.25 & 26.15 & 35.71 \\
\textsc{Qwen3-235B-A22B-Thinking} & adaptive & 13.49 & 20.82 & 13.34 & 21.11 \\
\textsc{DeepSeek-V3.2} & adaptive & 15.33 & 25.47 & 10.48 & 26.44 \\
\textsc{GLM-5} & adaptive & 18.92 & 37.17 & 28.87 & 44.25 \\
\textsc{GPT-5.4} & none & 9.41 & 20.96 & 15.38 & 26.54 \\
\textsc{GPT-5.4} & medium & 17.81 & 53.47 & 26.88 & 68.22 \\
\textsc{Gemini-3-Flash} & minimal & 24.99 & 41.10 & 39.98 & 54.34 \\
\textsc{Gemini-3-Flash} & medium & 35.08 & 52.16 & 51.82 & 67.25 \\
\textsc{Gemini-3.1-Pro} & low & 46.43 & 67.54 & 64.10 & 80.49 \\
\textsc{Gemini-3.1-Pro} & medium & 49.10 & 72.97 & 65.79 & 83.60 \\
\textsc{Gemini-3.1-Pro} & high & \textbf{51.29} & \textbf{76.03} & \textbf{67.78} & \textbf{85.74} \\
\bottomrule
\end{NiceTabular}
}
\caption{Overall performance results.}
\label{tab:main-results}
\end{table}

\paragraph{Task \& Evaluation Methods}
\label{sec:evaluation}
We formulate our problem as an open-ended question answering task. Since our benchmark requires substantial reasoning over real-world knowledge, we evaluate non-thinking models using Chain-of-thought prompting and thinking models using a standard prompt with varying levels of thinking effort.
We evaluate all models using their default recommended decoding setup. Evaluation prompts can be found in Appendix Table \ref{tab:prompt-evaluation}.

\paragraph{Evaluation Metrics}
\label{sec:metrics}
We measure the model performance using two metrics: \textbf{Exact Match Accuracy}, where we match the reference answer with the model response after light normalization (i.e., lowercase, remove punctuation, and unicode normalization for Russian), and \textbf{LLM-as-a-judge Accuracy}, where we employ \texttt{GPT-4o} to judge the correctness of the model response given the reference answer \citep{gu2024survey}. To ensure high-quality LLM judgment, we instruct it to ignore typos, articles, or formatting differences and provide it with additional context, including the human answer explanations and other acceptable answers, if any. LLM-as-a-judge prompt can be found in Appendix Table \ref{tab:prompt-llm-judge}.
\label{sec:experiments}

\section{Results}
\paragraph{Overall Performance}
Table~\ref{tab:main-results} reports overall performance on \ourbench{}-En and \ourbench{}-Ru. Results vary widely across models and metrics, ranging from below $10\%$ to above $80\%$. Non-thinking models all remain under $30\%$ accuracy, with \texttt{GPT-4.1} being the strongest in this group. Thinking models generally outperform their non-thinking counterparts, and greater thinking effort consistently yields higher accuracy. This effect is particularly pronounced for \texttt{GPT-5.4}, whose performance nearly doubles when moving from no thinking to medium effort. Among thinking models, closed-source models lead overall; the \texttt{Gemini} series in particular achieves top performance even at minimal thinking effort, surpassing all open-weight models.

\begin{figure*}[t]
\includegraphics[width=\linewidth, trim={0cm 0cm 0cm 0cm},clip]{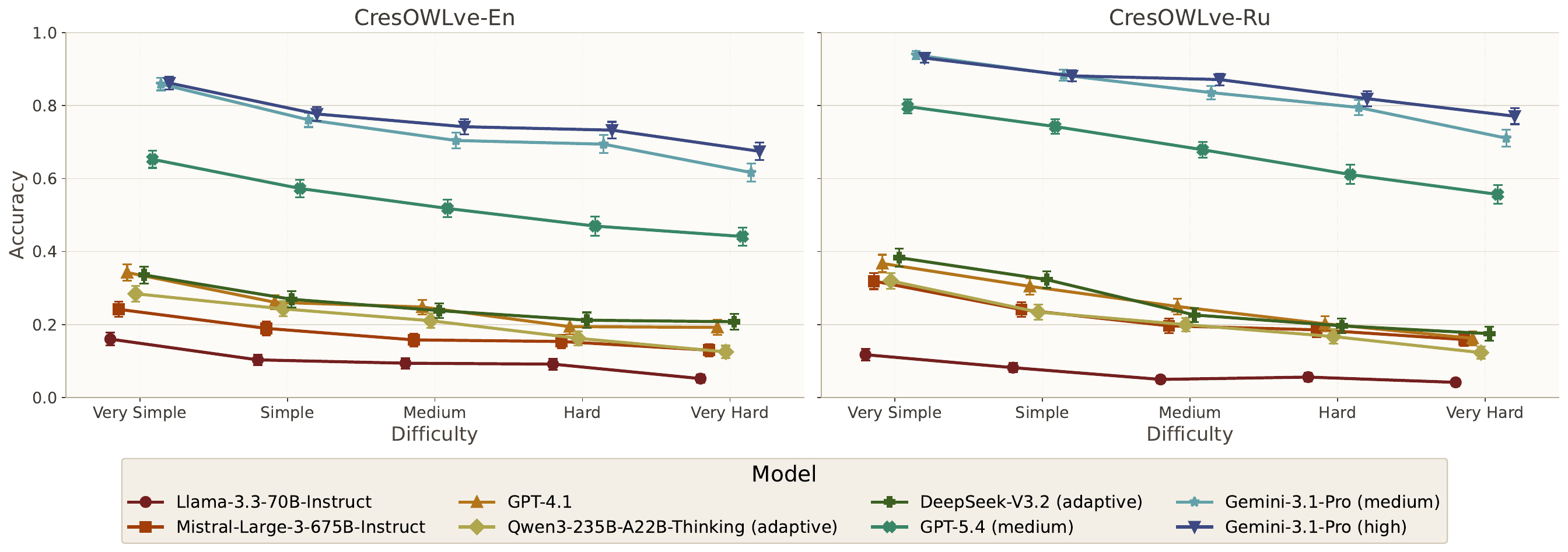}
\caption{LLM Judge results by difficulty level (Exact Match, Appendix Figure \ref{fig:diff-results-em}).}
\label{fig:diff-results-judge}
\end{figure*}

Regarding cross-lingual transfer, we observe mixed trends. Most non-thinking models suffer a notable performance drop on the Russian benchmark (up to $-5\%$ LLM Judge and $-3\%$ Exact Match), with \texttt{Qwen3-235B-A22B-Instruct} and \texttt{GPT-4.1} being the only exceptions. Thinking models, however, show the opposite trend: closed-source thinking models in particular achieve substantially higher scores on the Russian benchmark (up to $+15\%$ LLM Judge and $+18\%$ Exact Match). 
This surprising advantage on \ourbench{}-Ru may reflect stronger multilingual reasoning capabilities at higher levels of thinking, or potential data contamination (for which we find some evidence in Appendix \ref{sec:contamination}). Regardless, the consistently low performance of open-weight (and some proprietary) models across both languages confirms that \ourbench{} poses a genuine challenge and serves as a robust testbed for benchmarking creative problem-solving in LLMs.


\paragraph{Performance by difficulty level}
Figure~\ref{fig:diff-results-judge} illustrates model performance stratified by difficulty level for both \ourbench{}-En and \ourbench{}-Ru under LLM Judge evaluation. As expected, accuracy decreases monotonically with difficulty for all models, confirming that the benchmark difficulty levels capture the underlying complexity of the questions. Across both benchmarks, the performance degradation from Very Simple to Very Hard is substantial — even the strongest thinking models lose over $20\%$ accuracy. This highlights a key challenge posed by \ourbench{}: while average accuracy figures may appear moderate, the benchmark contains a non-trivial proportion of questions that remain genuinely difficult even for state-of-the-art models, underscoring the headroom for future progress.

We also provide detailed performance breakdowns by knowledge and creative domains, and culture groups in Appendix \ref{sec:performance-dom-cre-cult}.


\label{sec:results}

\section{Analysis}
\begin{figure*}[t]
\includegraphics[width=\linewidth, trim={0cm 0cm 0cm 0cm},clip]{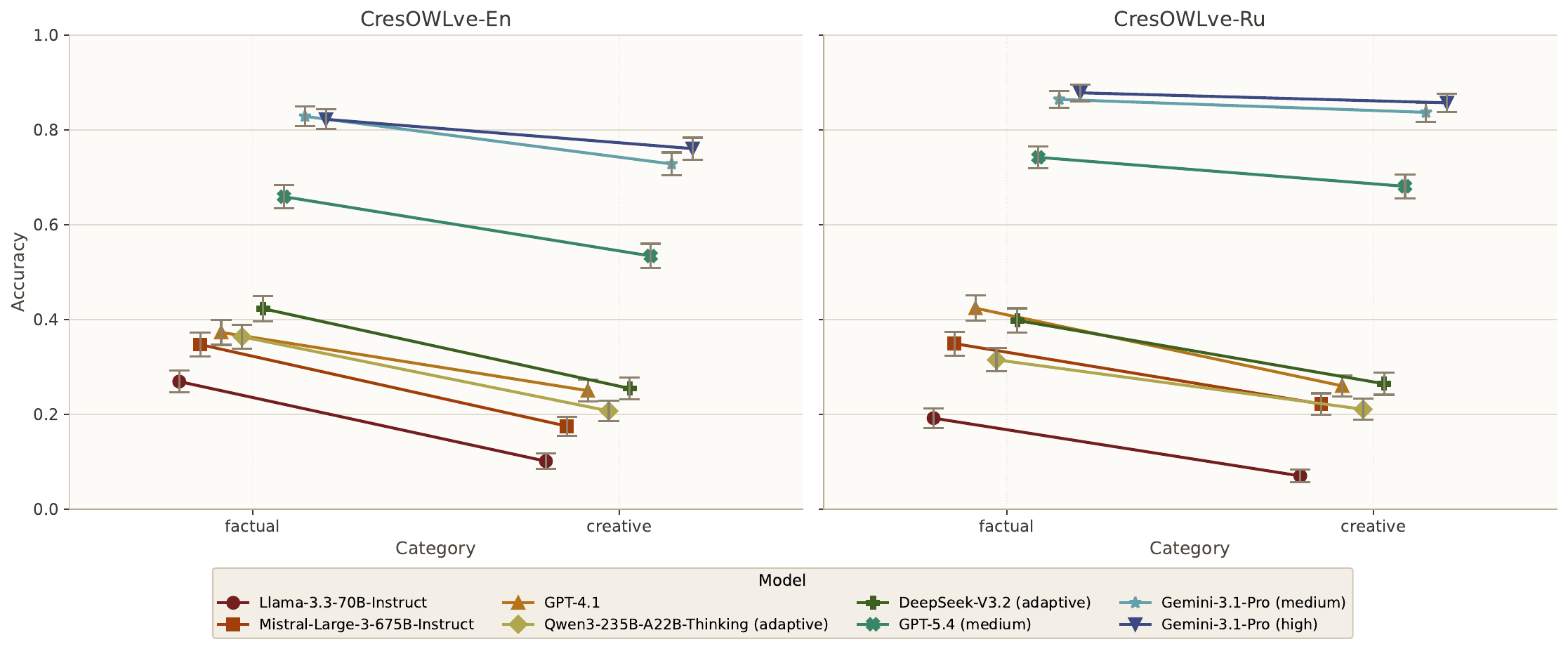}
\caption{LLM Judge results by reasoning category (Exact Match, Appendix Figure \ref{fig:cre-fac-results-em}). }
\label{fig:cre-fac-results-judge}
\end{figure*}

\paragraph{Creative vs. Factual Reasoning}
\label{sec:cre-factual-reasoning}
To investigate whether models struggle more with creative reasoning than factual retrieval, we evaluate on the factual subset of \ourbench{} as discussed in Section~\ref{sec:creative_vs_factual}. Figure \ref{fig:cre-fac-results-judge} reveals a consistent and substantial performance drop from factual to creative questions across all models and languages. 
On \ourbench{}-En, performance drop ranges from $-6.08\%$ for \texttt{Gemini-3.1-Pro (high)} to $-17.21\%$ for \texttt{Mistral-Large-3-675B-Instruct}. On \ourbench{}-Ru, the same trend holds, though the absolute drop is smaller for \texttt{Gemini-3.1-Pro} models ($-2.14\%$ and $-2.73\%$). For a fair comparison, we match the sample sizes across both categories and perform bootstrap sampling with 1000 iterations. In addition, we balance the number of samples across difficulty levels for both creative and factual questions (Appendix Figure \ref{fig:creative-factual-diff-dist}). These results confirm that creative questions pose a fundamentally harder challenge than factual ones.

\paragraph{Difficulty vs. Complexity}
\label{sec:diff-complexity}
To better understand the sources of question difficulty in \ourbench{}, we investigate whether difficulty correlates with measurable complexity features of the questions. Specifically, we consider three proxy measures of question complexity: the number of domains involved, the maximum semantic distance between domains for a given question, and the number of atomic facts required to resolve it. The latter is estimated by prompting \texttt{GPT-4o} to decompose each question into a list of constituent factual sub-questions (prompt provided in Appendix Table \ref{tab:prompt-knowledge}). Despite the intuitive appeal of these features as predictors of difficulty, we find no significant correlation between any of them and either the assigned difficulty level or model performance (Appendix Figures \ref{fig:diff-corrs}, \ref{fig:perf-corrs}). These findings strongly suggest that the difficulty of questions in \ourbench{} doesn't stem from the number of involved domains, the challenge of bridging semantically distant domains, or the volume of facts one needs to recall, but rather from the need to connect knowledge in creative ways.

\paragraph{Error Analysis}
\label{sec:error-analysis}
We conduct a manual analysis of errors produced by Gemini-3-Flash (medium). From a total of 1,087 error examples, we randomly sample 150 instances with the model's reasoning traces and obtain annotations from three authors of the paper, who are also domain experts. Following an iterative error analysis process, we identify seven error categories that are most prominent in our data:
\begin{itemize}
    \item \textbf{Missing creative connection:} The model retrieves relevant knowledge, but fails to recognize the intended associative or metaphorical link between clues, preventing it from making the key conceptual leap needed to reach the correct answer.
    \item \textbf{Overthinking:} The model identifies the correct concept during reasoning, but later replaces it with another answer due to reinterpretation or over-generalization.
    \item \textbf{Hallucination or unsupported fabrication:} The model invents facts, explanations, or source details that are not supported by the question or reliable knowledge.
    \item \textbf{Incorrect concept anchoring:} The model locks onto an incorrect concept early and builds reasoning around it, either through associative drift or by reinterpreting the clues to fit that concept.
    \item \textbf{Wrong hypothesis:} The model infers an incorrect rule or shared property from the clues and applies it consistently to produce an answer.
    \item \textbf{Wrong reference:} The model retrieves and reasons about an incorrect work, person, event, or source that superficially matches the clues but is not the intended reference.
    \item \textbf{Clue misinterpretation:} The model identifies the general source, context, or line of reasoning, but selects the wrong specific element required by the clue.
\end{itemize}

\begin{figure*}[t]
\includegraphics[width=\linewidth, trim={0cm 0cm 0cm 0cm},clip]{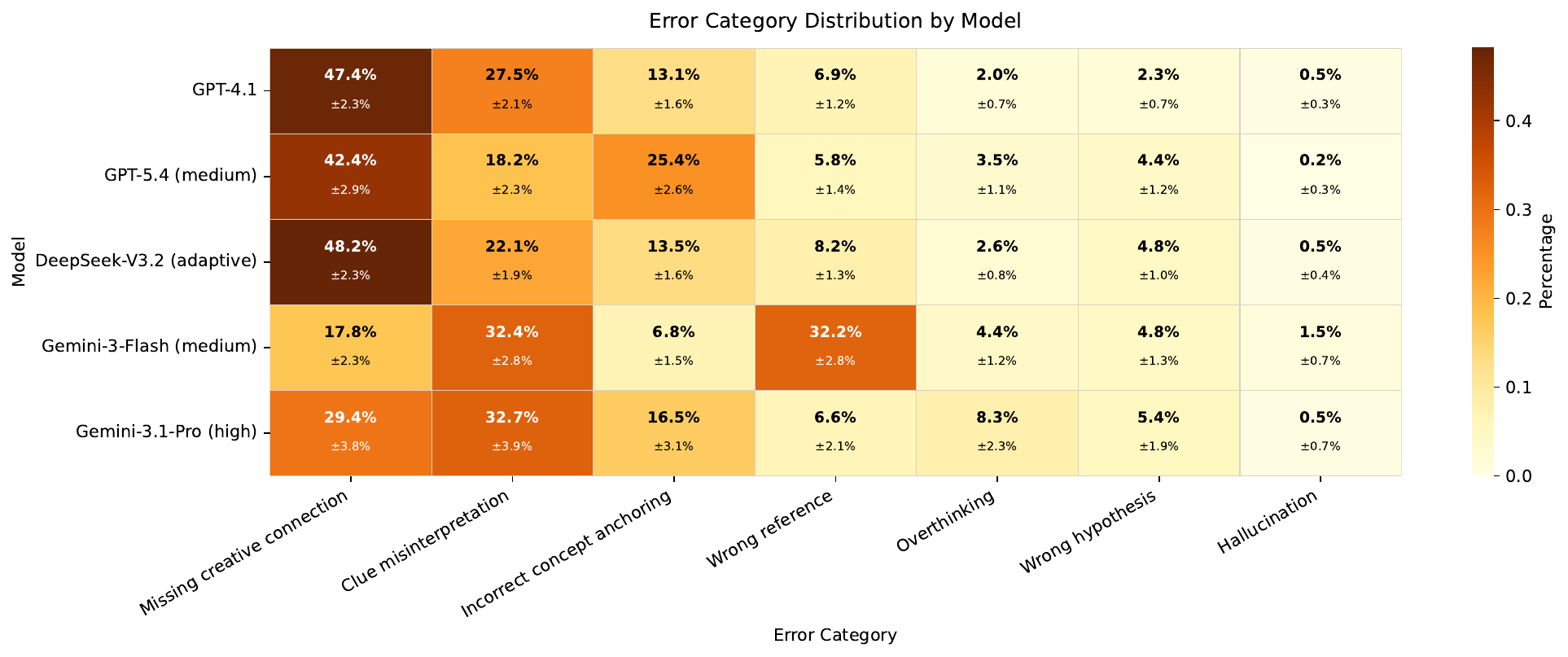}
\caption{Error category distribution for best-performing models (Wilson $95\%$ CI half-widths below each proportion, all within 1–3 percentage points).}
\label{fig:error-cat}
\end{figure*}

For each category, we provide representative examples in Appendix~\ref{app:error_ex}. We then use these definitions to automatically label all wrong examples ($N = 554-1,761$) at scale. 
The resulting error distribution for the best performing models is shown in Figure~\ref{fig:error-cat}. The most frequent error categories across models are \textit{missing creative connection} and \textit{clue misinterpretation}, further confirming our hypothesis about the difficulty of questions. A notable exception is \texttt{Gemini-3.1-Flash}, which shows significantly fewer missing creative connection errors but a substantially higher rate of \textit{wrong reference} errors. \texttt{GPT-5.4} stands out for its disproportionately high rate of \textit{incorrect concept anchoring}. Finally, low hallucination and wrong reference rates across most models confirm that failures stem not from missing knowledge but from an inability to make the creative connections between facts necessary to reach the correct answer.

\label{sec:analysis}

\section{Discussion}
\ourbench{} provides a bilingual benchmark for creative problem-solving over real-world knowledge, built from expert-authored ``What? Where? When?'' puzzles and validated through a multi-stage filtering pipeline. The benchmark remains highly challenging even for frontier thinking models, and our analysis attributes this difficulty not to surface complexity or knowledge retrieval but to the need to forge non-obvious creative connections between facts. We close by clarifying several aspects of the benchmark's design and scope.

\paragraph{Process creativity, not output creativity.} \ourbench{} evaluates the reasoning required to reach an answer rather than the novelty of the answer itself. Expected outputs are short, objective, and largely canonical, with limited surface variation; where alternative correct answers exist, they are enumerated and provided to the LLM judge alongside human explanations. The judge therefore only verifies whether a response matches a known answer modulo formatting, mitigating the risk of penalizing unconventional valid responses.

\paragraph{Creative connection, not multi-hop retrieval.} Three lines of evidence indicate the difficulty reflects creative reasoning rather than longer retrieval chains. First, neither difficulty rating nor accuracy correlates with surface complexity features (number of domains, semantic distance, atomic facts; Appendix Figures \ref{fig:diff-corrs}, \ref{fig:perf-corrs}). Second, error analysis (\S\ref{sec:analysis}) shows missing creative connection and clue misinterpretation dominate failures, while hallucinations and wrong-reference errors (signatures of retrieval failure) are rare. Third, the factual subset matches the creative subset on difficulty and retrieval demand (Appendix Figure \ref{fig:creative-factual-diff-dist}) but lacks the associative leap; the consistent factual-vs-creative gap (up to $-17\%$) isolates creative connection as the source of difficulty.

\paragraph{Data contamination.} As with any benchmark from publicly available material, contamination is a shared concern \citep{xu2024benchmark}, and contamination-free corpora of comparable scale are prohibitively costly for subjective creative content. \ourbench{} items were nonetheless authored by human experts over decades, mostly predating the LLM era, ensuring quality free from AI-generated artifacts. We include a contamination analysis (Appendix \ref{sec:contamination}) to make the memorization-vs-reasoning distinction transparent. While it suggests memorization plays a role for \texttt{Gemini-3.1-Pro} on the Russian subset, our conclusions are not driven by this: reasoning effort scales performance monotonically within a fixed model (\texttt{GPT-5.4} nearly doubles from no thinking to medium effort, $9.4\%$→$17.8\%$ En, $15.4\%$→$26.9\%$ Ru), and the thinking advantage holds for open-weight models (\texttt{DeepSeek-V3.2}, \texttt{GLM-5}, \texttt{Qwen3.5-397B}) where the cross-lingual asymmetry is absent. Most models remain below $30\%$ accuracy, and the factual-vs-creative gap holds throughout, indicating creative reasoning is the dominant bottleneck.

\paragraph{Human performance.} We do not conduct a formal human evaluation, as \ourbench{} is not a human-LLM comparison benchmark; it measures creative problem-solving where the relevant knowledge should already be accessible to the model. Every item was correctly answered at least once by expert teams in the original game, an implicit $100\%$ human-solvability upper bound. The conditions are not directly comparable, however: humans play live, time-constrained, and without internet access, while LLMs train on effectively the entire web. That substantial gaps persist despite this asymmetry underscores the paper's central finding: humans solve from partial knowledge through creativity, while LLMs with broader recall still fall short.

\label{sec:discussion}

\section{Conclusion}
We introduced \ourbench{}, a bilingual benchmark for evaluating creative problem-solving in LLMs, constructed from real-world puzzles sourced from the What?\ Where?\ When?\ intellectual game. Our evaluation of several frontier LLMs reveals that the benchmark remains highly challenging, with even the strongest thinking models falling considerably short on creative questions. Analysis shows that this difficulty does not stem from surface-level complexity features, but from the need to forge creative connections between knowledge from different domains. We hope \ourbench{} serves as a challenging testbed to drive progress in creative reasoning.
\label{sec:conclusion}

\section*{Ethics Statement}
This work introduces \ourbench{}, a benchmark for evaluating creative problem-solving in large language models using puzzles derived from the intellectual game ``What? Where? When?''. 
The benchmark is constructed from publicly available questions created by human authors over several decades. These questions reflect the intellectual contributions of their original creators, and we do not claim authorship of the underlying content. The dataset is used solely for research purposes, and proper attribution should be maintained where applicable.

Because the questions are designed to test non-obvious associative and creative connections, they may reflect the cultural context, assumptions, and potential biases of their authors. In particular, as the source material originates in Russian, some questions may encode culturally specific knowledge or perspectives. To mitigate this, we annotate questions based on their regional content; however, one should account for potential biases.
Given that the source material is publicly available and widely distributed, there is a risk that some models may have been exposed to similar questions during training. As discussed in our analysis, this potential data contamination may affect performance and should be considered when interpreting results. 

Finally, we believe \ourbench{} makes a positive contribution to the NLP research community by providing a challenging evaluation resource that goes beyond factual recall and isolated reasoning skills. By grounding evaluation in real-world puzzles that require creative synthesis of knowledge, we hope to encourage the development of LLMs with more robust and flexible reasoning capabilities. At the same time, we caution against using benchmark performance as a sole measure of creative intelligence, as creativity is inherently difficult to define and evaluate. Our categorization into factual and creative questions, as well as the proposed error taxonomy, involves subjective judgments and may not capture all aspects of creative reasoning.
\label{sec:ethics}

\section*{Acknowledgements}
We thank the members of the EPFL NLP for their feedback on the project and the paper manuscript. We gratefully acknowledge the support of the Swiss National Science Foundation (grant 205121\_207437: C - LING) and Fondazione Aldo e Cele Daccò. AB also gratefully acknowledges the support of the Swiss National Science Foundation (No. 215390), the European Research Council (Starting grant no. 101222478, RESPECT-LM), the AI2050 program at Schmidt Sciences (Grant \#G-25-69783), Sony Group Corporation, and the Swiss National Supercomputing Center (CSCS) in the form of an infrastructure engineering and development project. 
\label{sec:acknowledgements}

\bibliography{colm2026_conference}
\bibliographystyle{colm2026_conference}

\newpage

\appendix

\section{Data Contamination Analysis}
\label{sec:contamination}
To investigate whether the performance advantage of thinking models on \ourbench{}-Ru over \ourbench{}-En could be attributed to memorization and potential data contamination, we conduct a targeted contamination analysis. To isolate other confounding factors, we restrict our analysis to the 580 questions that \texttt{Gemini-3.1-Pro} answered correctly in Russian but not in English. We focus on this model as the best-performing one in our evaluation, and deliberately use low thinking effort to ensure the model relies more on memorization than active reasoning. Following \citet{golchin2025data}, we first generate three unique word-level perturbations for each question using \texttt{GPT-4o} (prompt in Table \ref{tab:prompt-contamination-perturb}). We then run two contamination quizzes sharing the same prompt structure (prompt in Table \ref{tab:prompt-contamination-quiz}) but with different answer choices. In the first quiz, we present the model with a multiple-choice question where all options are perturbed versions of the original question and an additional ``none of the above'' option is included; crucially, no correct answer is present. This allows us to identify positional biases by analyzing how frequently the model selects each option. We identify non-preferred positions as those chosen less frequently than random chance (less than $k/4=145$ with $k=580$ and 4 options), which yields options B and C as non-preferred(full distribution is {'A': 187, 'B': 60, 'C': 47, 'D': 273} with an additional 13 invalid answers). Notably, the model correctly prefers option D (``none of the above'') most of the time. In the second quiz, we insert the correct answer into one of these non-preferred positions (randomly chosen between B and C) for each question and repeat this quiz three times. The model achieves $72\%$, $73\%$, and $74\%$ accuracy across the three runs (mean $73\%$), all substantially above the random baseline of $25\%$. This result strongly suggests that \texttt{Gemini-3.1-Pro} has likely been exposed to these questions during training, pointing to high-level data contamination as a plausible explanation for its cross-lingual performance advantage.

\section{Annotation Reliability}
\label{sec:annot-reliability}
To assess annotation reliability, two of the authors independently re-annotated a sample of $50$ items along the four critical fields, which we annotated using \texttt{GPT-4o}: reasoning type (factual/creative), creative concepts, knowledge domains, and cultures. Then we computed the inter-annotator agreements between the humans and the model:
\paragraph{Reasoning type (single-label binary)} Raw agreement = $0.77$, Fleiss $\kappa = 0.55$ \citep{fleiss1971measuring} - moderate agreement, with disagreement concentrated on borderline factual/creative items, consistent with the inherent subjectivity we note in our Ethics Statement.
\paragraph{Multi-label fields} We use Gwet's AC1 \citep{gwet2014handbook}, support-weighted across labels; AC1 is used due to high label-prevalence skew that deflates Fleiss $\kappa$:
\begin{itemize}
    \item \textbf{Creative concepts:} AC1 = $0.79$ (top labels — metaphor: $0.91$, analogy: $0.86$, pun: $0.87$, poem: $0.97$, idiom: $0.99$; the broader lateral thinking label is noisier at $0.42$, reflecting its function as a general descriptor in our taxonomy).
    \item \textbf{Knowledge domains:} AC1 = $0.85$ (e.g., Astronomy: $1.00$, Mathematics: $0.99$, Literature: $0.85$; History is the main soft spot at $0.53$, where annotators differ on whether to tag historical figures or only historical events).
    \item \textbf{Cultures:} AC1 = $0.90$ (e.g., Russian: $0.82$, German: $0.97$, Italian: $1.00$).
\end{itemize}

\begin{table*}[h]
\centering
\small
\scalebox{0.9}{
\begin{tabular}{lcccc}
\toprule
Difficulty & \#questions & \#explanations & \makecell{Avg. \#question tokens \\ (En / Ru)} & \makecell{Avg. \# answer tokens \\ (En / Ru)} \\
\midrule
Very Simple (1) & $417$ & $375$ & $37.79\pm12.84$ / $29.65\pm9.90$ & $2.12\pm1.51$ / $1.82\pm1.09$ \\
Simple (2) & $453$ & $426$ & $38.64\pm12.84$ / $30.79\pm10.13$ & $2.20\pm1.73$ / $1.87\pm1.30$ \\
Medium (3) & $436$ & $413$ & $38.11\pm12.13$ / $30.34\pm10.08$ & $2.06\pm1.38$ / $1.87\pm1.10$ \\
Hard (4) & $371$ & $355$ & $38.61\pm13.43$ / $30.60\pm10.35$ & $2.13\pm1.88$ / $1.85\pm1.39$ \\
Very Hard (5) & $384$ & $359$ & $37.92\pm13.05$ / $30.27\pm10.45$ & $2.12\pm1.46$ / $1.89\pm1.22$ \\
\midrule
\textbf{Total} & $\mathbf{2061}$ & $\mathbf{1928}$ & $\mathbf{38.22\pm12.84}$ / $\mathbf{30.33\pm10.18}$ & $\mathbf{2.13\pm1.60}$ / $\mathbf{1.86\pm1.22}$ \\
\bottomrule
\end{tabular}
}
\caption{Final benchmark statistics by difficulty level.}
\label{tab:data-stats}
\end{table*}

\section{Performance by domain, creativity, and cultures}
\label{sec:performance-dom-cre-cult}
\begin{figure}[h]
\includegraphics[width=\linewidth, trim={0cm 0cm 0cm 0cm},clip]{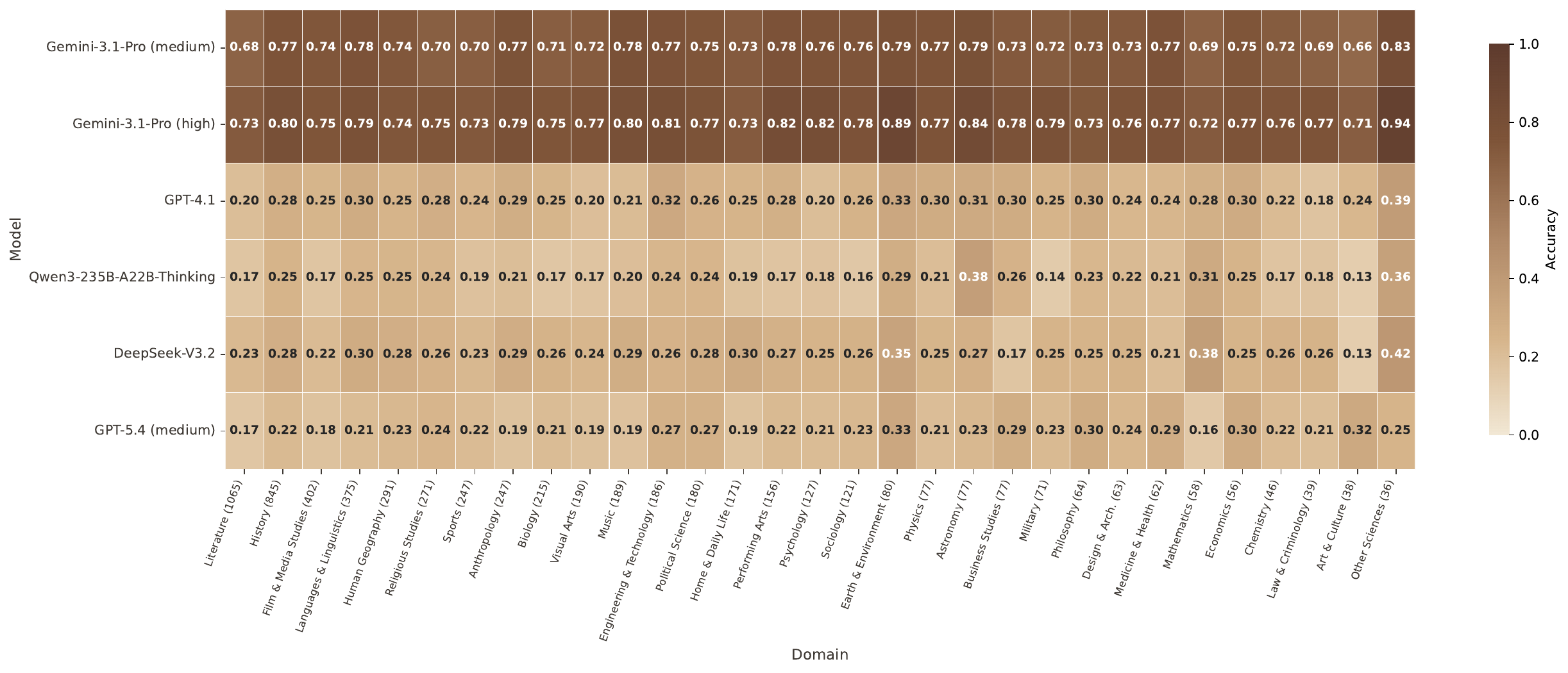}
\caption{Performance by domains on \ourbench{}-En.}
\label{fig:domain-results-en}
\end{figure}

\begin{figure}[h]
\includegraphics[width=\linewidth, trim={0cm 0cm 0cm 0cm},clip]{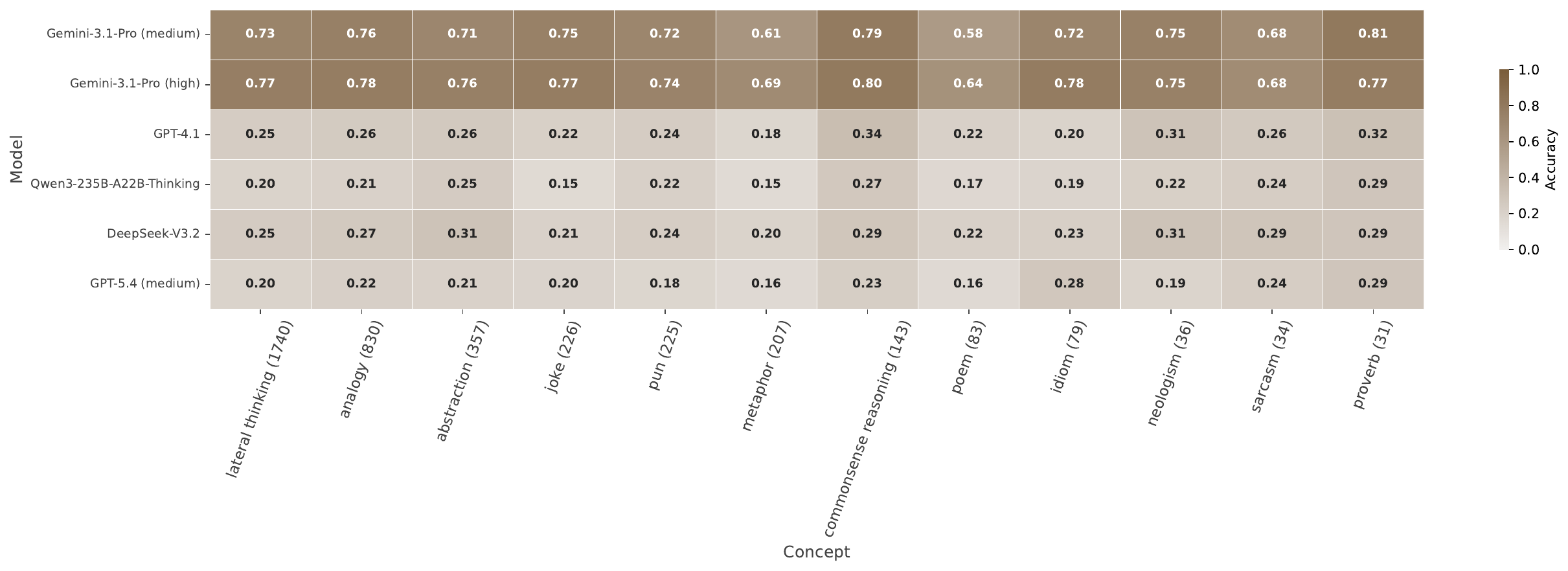}
\caption{Performance by creativity concepts on \ourbench{}-En.}
\label{fig:creative-results-en}
\end{figure}

\begin{figure}[h]
\includegraphics[width=\linewidth, trim={0cm 0cm 0cm 0cm},clip]{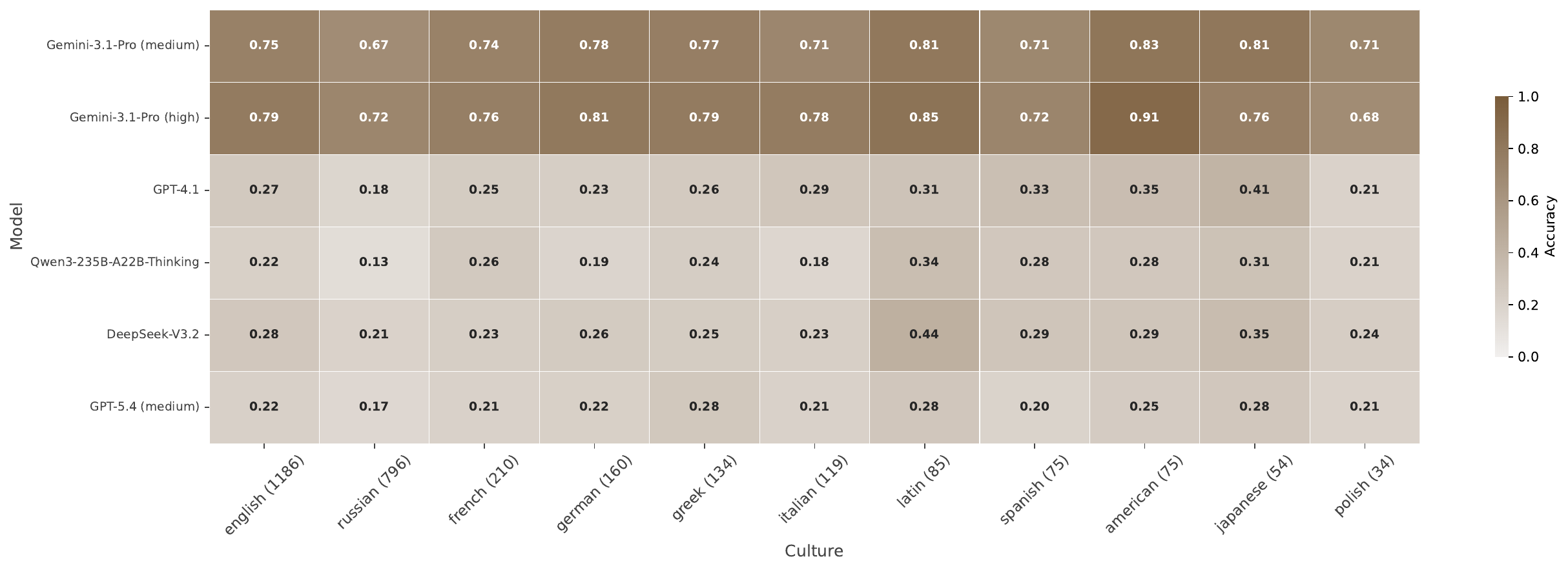}
\caption{Performance by cultures on \ourbench{}-En.}
\label{fig:culture-results-en}
\end{figure}

\begin{figure}[h]
\includegraphics[width=\linewidth, trim={0cm 0cm 0cm 0cm},clip]{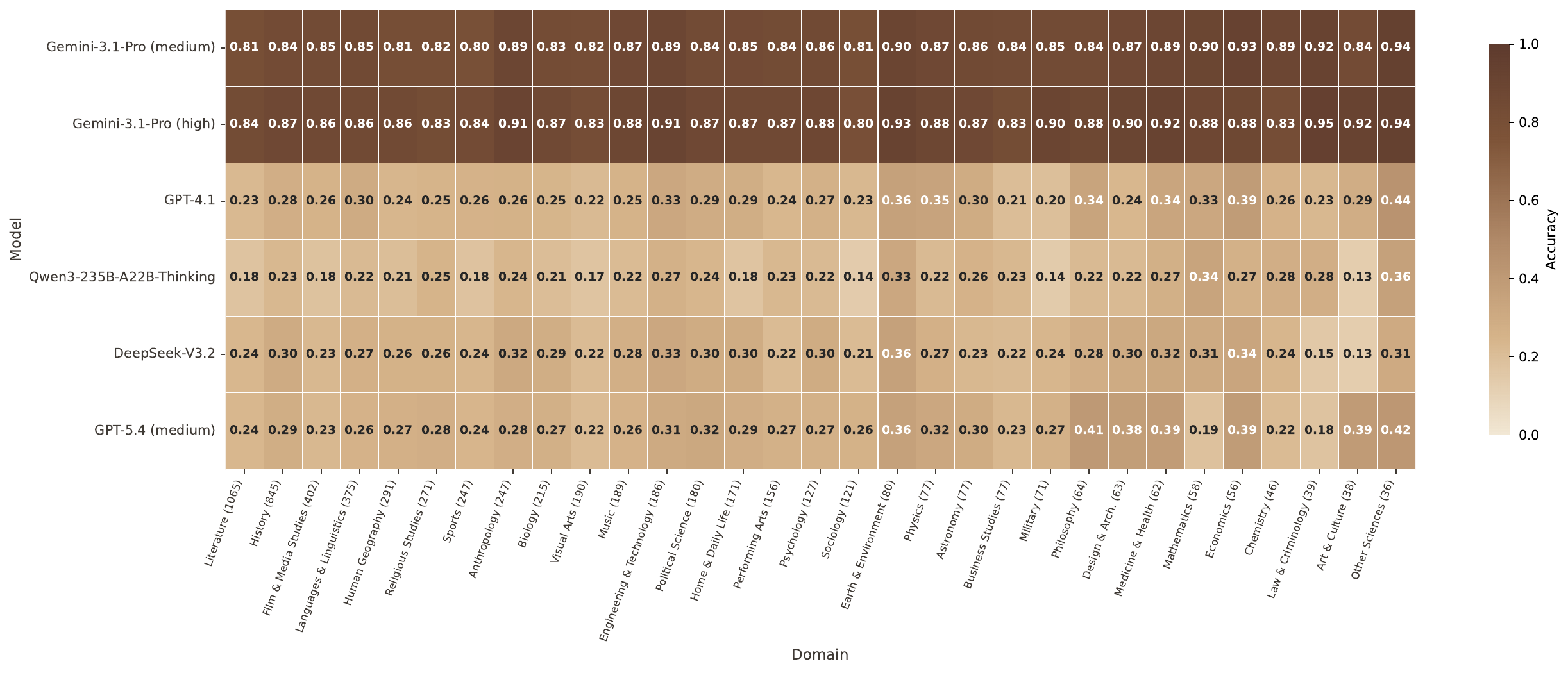}
\caption{Performance by domains on \ourbench{}-Ru.}
\label{fig:domain-results-ru}
\end{figure}

\begin{figure}[h]
\includegraphics[width=\linewidth, trim={0cm 0cm 0cm 0cm},clip]{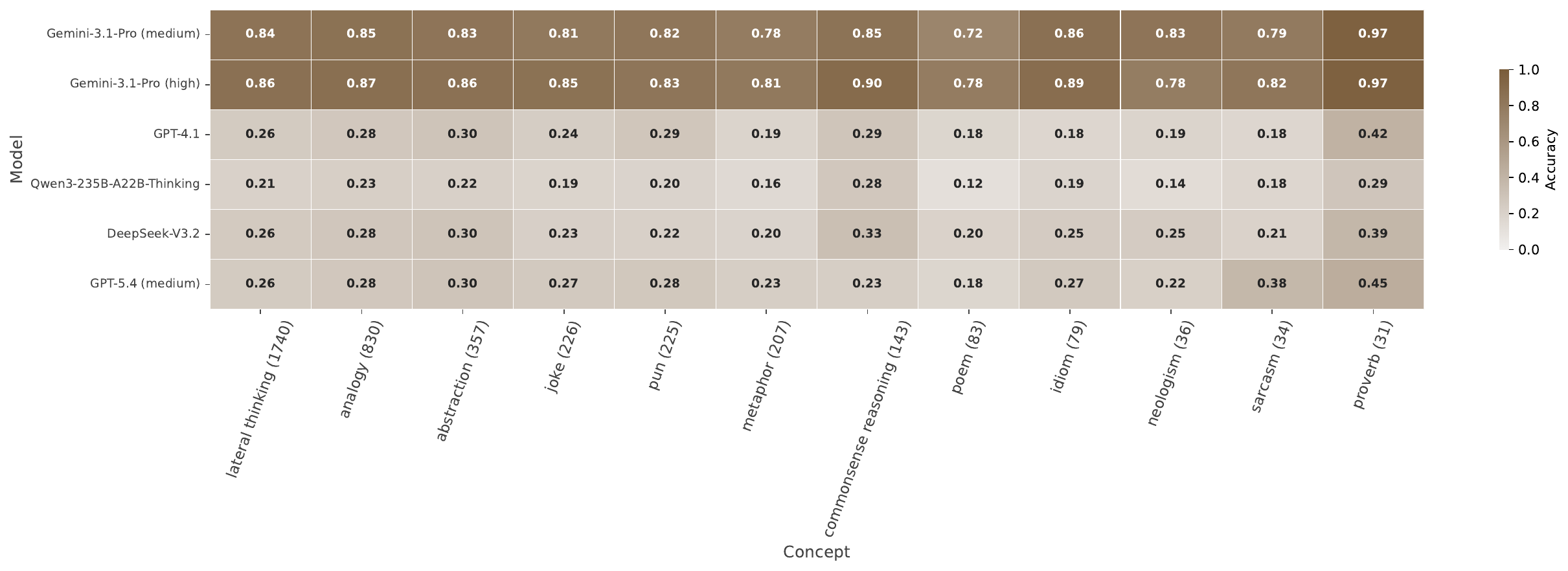}
\caption{Performance by creativity concepts on \ourbench{}-Ru.}
\label{fig:creative-results-ru}
\end{figure}

\begin{figure}[h]
\includegraphics[width=\linewidth, trim={0cm 0cm 0cm 0cm},clip]{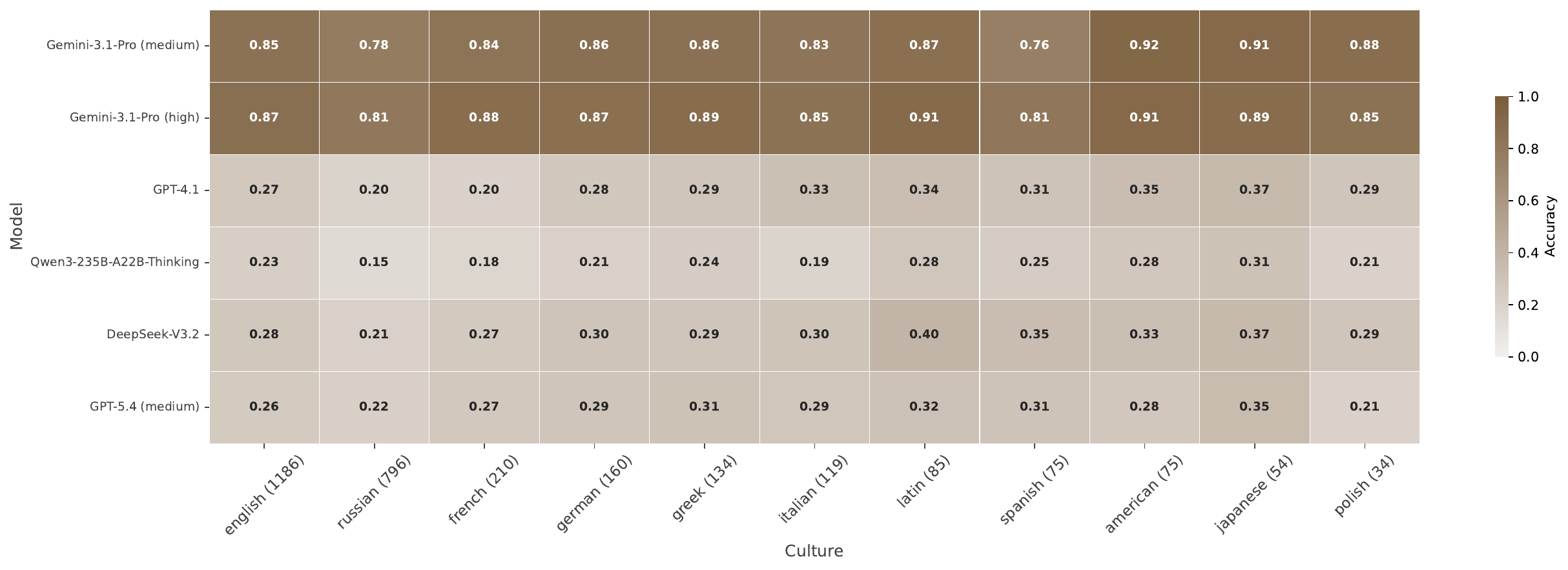}
\caption{Performance by cultures on \ourbench{}-Ru.}
\label{fig:culture-results-ru}
\end{figure}

Detailed performance breakdowns by knowledge and creative domains, and culture groups are provided in Figures \ref{fig:domain-results-en}, \ref{fig:creative-results-en}, \ref{fig:culture-results-en} and \ref{fig:domain-results-ru}, \ref{fig:creative-results-ru}, \ref{fig:culture-results-ru} for the \ourbench{}-En and \ourbench{}-Ru benchmarks respectively. Across benchmarks, performance is largely uniform across all three stratifications, though several notable trends emerge. Across domains, most models tend to perform better on Earth \& Environmental Sciences questions, with \texttt{Qwen3-235B-A22B-Thinking} showing a particular advantage on Astronomy and \texttt{DeepSeek-V3.2} on Mathematics. Regarding creative language concepts, models generally perform worse on questions involving poems and metaphors, suggesting that figurative and poetic language remains a particular challenge. On the culture dimension, \texttt{Gemini-3.1-Pro} performs notably well on questions involving American cultural references, \texttt{GPT-4.1} on Japanese, and \texttt{DeepSeek-V3.2} on Latin.

\section{Synthetic Benchmark Construction}
\label{sec:synthetic}

While \ourbench{} is sourced from expert-authored puzzles, the underlying construction principle — combining real-world facts through non-obvious creative connections — can in principle be replicated synthetically using structured knowledge sources. Here we describe a prototype pipeline that constructs \ourbench{}-style puzzles from a real-world knowledge graph, together with a preliminary feasibility study. We view this as a complementary, scalable route to generating creative-reasoning puzzles and a foundation for future work.

\paragraph{Pipeline.} Our pipeline proceeds in five stages:
\begin{enumerate}
\item \textbf{Anchor sampling.} We randomly sample a moderately-notable answer entity from Wikidata via SPARQL, restricted to a target domain (e.g. literature) and a notability band based on the number of sitelinks (between 20 and 100). This band selects entities that are identifiable from their factual descriptions yet not so iconic that they can be trivially recognized from indirect clues alone.
\item \textbf{Fact extraction.} We extract a small set of peripheral facts about the anchor entity (e.g. awards received, place of education) from its knowledge-graph neighborhood, favoring facts that jointly but not individually identify the entity.
\item \textbf{Verbalization.} We use an LLM verbalizer to produce two versions of the puzzle: a \textit{literal} version that states the extracted facts directly, and an \textit{obfuscated} version that re-expresses the same facts indirectly through metaphor, analogy, or abstraction, in the style of \ourbench{}.
\item \textbf{Creative-leap filtering.} We apply a strong solver to both versions and retain only items the solver answers correctly given the literal version but fails to answer given the obfuscated version. This gap isolates the \textit{creative connection} requirement: the solver possesses the underlying knowledge (evidenced by solving the literal version) but cannot make the non-obvious associative leap required by the obfuscated version — precisely the failure mode our benchmark targets (§\ref{sec:analysis}).
\item \textbf{Human validation.} Surviving items are passed to human validators to confirm answerability and remove ambiguous cases. We treat this stage as necessary for a finalized dataset but out of scope for the present feasibility study.
\end{enumerate}

\paragraph{Feasibility study.} We implemented a prototype of this pipeline using \texttt{Gemini-3-Flash-Preview} as both verbalizer and solver, sampling anchor entities from the literature domain. Of 100 candidate entities, 7 survived the creative-leap filter — a non-trivial yield given that all stages must succeed jointly, and one that could be improved through verbalizer prompt tuning and broader anchor sampling. Table~\ref{tab:synth_example} shows a representative surviving example.

\paragraph{Extensions.} The prototype admits several natural extensions we leave for future work. First, the binary correctness criterion can be relaxed to admit \textit{multiple valid answers}: an obfuscated puzzle may be legitimately satisfiable by more than one entity (e.g. distinct authors matching the same indirect description), and enumerating these would both improve evaluation fairness and enrich the benchmark. Second, obfuscations vary in difficulty, and \textit{automated difficulty ratings} could be assigned based on the semantic or embedding distance between the literal facts and their obfuscated rendering. Beyond calibrating difficulty, this would enable a new axis of analysis: on instances admitting multiple valid answers of varying association distance, one could probe whether models systematically favor closer, more accessible associations over distant, more creative ones. Together with human validation at scale, these extensions would turn the prototype into a fully synthetic, scalable complement to \ourbench{}.

\begin{table}[h]
\centering
\small
\begin{tabular}{p{0.15\textwidth}p{0.78\textwidth}}
\toprule
\textbf{Anchor entity} & Günter Wallraff \\
\midrule
\textbf{Facts} & received the award (P166) $\rightarrow$ ``Gerty Spies Award''; received the award (P166) $\rightarrow$ ``August Bebel Prize'' \\
\midrule
\textbf{Literal} & X received the Gerty Spies Award and the August Bebel Prize. Who is X? \newline \textit{Solver: Günter Wallraff} \Checkmark \\ \midrule \textbf{Obfuscated} & To unmask the world, X masks himself. His labor has been crowned with the names of a poet who endured the camps and a man who forged the hammer of the people. Who is this master of assumed identities? \newline \textit{Solver: Subcomandante Marcos} \XSolidBrush \\
\midrule
\textbf{Explanation} & Günter Wallraff is an undercover journalist (``masks himself''); Gerty Spies is a poet and Holocaust survivor (``endured the camps''); August Bebel is a socialist politician (``forged the hammer of the people''). The solver retrieves related but incorrect entities, failing to make the intended connections. \\
\bottomrule
\end{tabular}
\caption{A representative example produced by the synthetic construction pipeline. The solver answers the literal version correctly but fails the obfuscated version, illustrating the creative-connection gap the pipeline is designed to elicit.}
\label{tab:synth_example}
\end{table}

\section{Error Examples}
\label{app:error_ex}

\begin{tcolorbox}[
  colback=white,
  colframe=bordergray,
  boxrule=0.5pt,
  arc=2mm,
  left=2mm,right=2mm,top=1mm,bottom=1mm,
  title=\textbf{Missing creative connection}
]

\textbf{Question}
\begin{tcolorbox}[
  colback=lightbluebg,
  colframe=darkblue,
  boxrule=0.4pt,
  arc=1.5mm,
  left=1.5mm,right=1.5mm,top=1mm,bottom=1mm
]
In the first half of the 20th century, many American farmers had to move---some due to economic difficulties, others due to dust storms. For example, \textbf{SHE} moved to another country in 1911. Name \textbf{HER} name.
\end{tcolorbox}

\textbf{Gold answer}
\begin{tcolorbox}[
  colback=lightgreenbg,
  colframe=darkgreen,
  boxrule=0.4pt,
  arc=1.5mm,
  left=1.5mm,right=1.5mm,top=1mm,bottom=1mm
]
Dorothy.
\end{tcolorbox}

\textbf{Model output}
\begin{tcolorbox}[
  colback=lightredbg,
  colframe=darkred,
  boxrule=0.4pt,
  arc=1.5mm,
  left=1.5mm,right=1.5mm,top=1mm,bottom=1mm,
  breakable
]
In the first half of the 20th century, a significant migration of American farmers occurred, often driven by economic opportunities and agricultural developments. One of the most notable periods of this migration was between 1896 and 1914, when hundreds of thousands of American farmers moved to the Canadian Prairies (Alberta, Saskatchewan, and Manitoba) to take advantage of cheap, fertile land. This period is historically recognized as the ``Great Migration'' of American farmers to Canada. In 1911 alone, over 100{,}000 Americans emigrated to Canada.

The placeholder \textbf{SHE} in the puzzle refers to a personification of the American farmer or a symbolic figure of the United States. Following the logic that SHE represents the collective identity of these migrants or a specific personification often used for ships or entities, the puzzle likely refers to the ship \textbf{``The American Farmer.''} However, a more direct interpretation of the riddle, considering the symbolic ``SHE'' often used for ships that transported these families, or the name of a famous individual, may point towards a specific ship or a personified noun. Yet, in the context of lateral thinking puzzles with these clues, the answer often points to a vessel or a specific entity named ``The American Farmer'' that ``moved'' (in maritime terms) to another country (destination) during that era. Looking at ship records, the SS \textit{American Farmer} was a famous ship, though it was renamed later.

Alternatively, some lateral thinking puzzles use ``SHE'' to refer to the \textbf{``Mona Lisa''}, which was stolen in \textbf{1911} and ``moved'' to Italy (another country). However, the connection to American farmers in the first sentence is typically a contextual setup for migration patterns of that era. Re-evaluating the direct clue, ``SHE'' is an example of an American farmer. In 1911, the vast number of farmers moving to Canada included many individuals, but the ``SHE'' most famously associated with the concept of the American farmer (in a figurative sense through art or iconography) or simply the ship named \textit{American Farmer} fits the phrasing. In maritime culture, a ship named \textit{American Farmer} would be referred to as ``SHE.''

\texttt{<Answer>The American Farmer</Answer>}
\end{tcolorbox}

\textbf{Predicted answer:} \textcolor{darkred}{The American Farmer}

\end{tcolorbox}

\begin{tcolorbox}[
  colback=white,
  colframe=bordergray,
  boxrule=0.5pt,
  arc=2mm,
  left=2mm,right=2mm,top=1mm,bottom=1mm,
  title=\textbf{Overthinking}
]

\textbf{Question}
\begin{tcolorbox}[
  colback=lightbluebg,
  colframe=darkblue,
  boxrule=0.4pt,
  arc=1.5mm,
  left=1.5mm,right=1.5mm,top=1mm,bottom=1mm
]
Pyotr Vail writes about one of the properties of cinema, that essentially D'Artagnan is no better than Rochefort, but \textbf{IT} convinces the viewer of D'Artagnan's rightness. ``Barbarossa'' was also \textbf{IT}. Name \textbf{IT} with one word.
\end{tcolorbox}

\textbf{Gold answer}
\begin{tcolorbox}[
  colback=lightgreenbg,
  colframe=darkgreen,
  boxrule=0.4pt,
  arc=1.5mm,
  left=1.5mm,right=1.5mm,top=1mm,bottom=1mm
]
Close-up.
\end{tcolorbox}

\textbf{Model output}
\begin{tcolorbox}[
  colback=lightredbg,
  colframe=darkred,
  boxrule=0.4pt,
  arc=1.5mm,
  left=1.5mm,right=1.5mm,top=1mm,bottom=1mm,
  breakable
]
To solve this lateral thinking puzzle, we look to the writings of \textbf{Pyotr Vail}, a renowned Russian-American interviewer and essayist. In his cultural commentary, Vail noted that in cinema, the protagonist (like \textbf{D'Artagnan} in \textit{The Three Musketeers}) is often no more morally upright than the antagonist (\textbf{Rochefort}); however, cinema uses specific techniques to make the viewer empathize with the hero.

The specific property Vail cites is the \textbf{``close-up''} (or in Russian, \textbf{``\foreignlanguage{russian}{крупный план}''}). By focusing the camera closely on the hero, cinema creates an intimate connection that ``convinces the viewer of the hero's rightness.''

The second clue refers to \textbf{``Barbarossa.''} In a historical and military context, ``Barbarossa'' was the name of a famous \textbf{plan} (the German invasion of the Soviet Union, known as \textit{Operation Barbarossa}).

In Russian, the word \textbf{``\foreignlanguage{russian}{план}''} (plan) can refer both to a military plan and to a cinematic shot scale (as in \textit{\foreignlanguage{russian}{крупный план}}). Based on this ambiguity, the model concludes that the correct answer is \textbf{plan}.

\texttt{<Answer>Plan</Answer>}
\end{tcolorbox}

\textbf{Predicted answer:} \textcolor{darkred}{Plan}

\end{tcolorbox}

\begin{tcolorbox}[
  colback=white,
  colframe=bordergray,
  boxrule=0.5pt,
  arc=2mm,
  left=2mm,right=2mm,top=1mm,bottom=1mm,
  title=\textbf{Hallucination or unsupported fabrication}
]

\textbf{Question}
\begin{tcolorbox}[
  colback=lightbluebg,
  colframe=darkblue,
  boxrule=0.4pt,
  arc=1.5mm,
  left=1.5mm,right=1.5mm,top=1mm,bottom=1mm
]
Upon learning about George Bush's upcoming visit to Israel, a representative of ``Al-Qaeda'' called him one of \textbf{THEM} and urged to meet Bush with weapons in hand. Name the writer who made \textbf{THEM} his main characters.
\end{tcolorbox}

\textbf{Gold answer}
\begin{tcolorbox}[
  colback=lightgreenbg,
  colframe=darkgreen,
  boxrule=0.4pt,
  arc=1.5mm,
  left=1.5mm,right=1.5mm,top=1mm,bottom=1mm
]
Henryk Sienkiewicz.
\end{tcolorbox}

\textbf{Model output}
\begin{tcolorbox}[
  colback=lightredbg,
  colframe=darkred,
  boxrule=0.4pt,
  arc=1.5mm,
  left=1.5mm,right=1.5mm,top=1mm,bottom=1mm,
  breakable
]
In the context of the puzzle, ``Al-Qaeda'' is interpreted as a pun referring to a group of alcoholics (phonetically similar to ``Al-Kashi'' in Russian jokes). The model assumes that George Bush, as a recovering alcoholic, is being described as one of ``THEM.'' It further interprets ``weapons'' as bottles and concludes that the question refers to literary works centered on alcoholics.

Based on this reasoning, the model identifies \textbf{Venedikt Yerofeyev}, known for \textit{Moscow-Petushki}, as the correct answer.

\texttt{<Answer>Venedikt Yerofeyev</Answer>}
\end{tcolorbox}

\textbf{Predicted answer:} \textcolor{darkred}{Venedikt Yerofeyev}

\end{tcolorbox}

\begin{tcolorbox}[
  colback=white,
  colframe=bordergray,
  boxrule=0.5pt,
  arc=2mm,
  left=2mm,right=2mm,top=1mm,bottom=1mm,
  title=\textbf{Hallucination or unsupported fabrication}
]

\textbf{Question}
\begin{tcolorbox}[
  colback=lightbluebg,
  colframe=darkblue,
  boxrule=0.4pt,
  arc=1.5mm,
  left=1.5mm,right=1.5mm,top=1mm,bottom=1mm
]
In Paris, Thomas Jefferson was presented with a gift made by French craftsmen from African materials. The gift has survived to this day, although not completely: two elements are missing. Perhaps they were not lost but deliberately removed to avoid putting the official representative of the young republican state in an awkward position. Name these two elements as precisely as possible.
\end{tcolorbox}

\textbf{Gold answer}
\begin{tcolorbox}[
  colback=lightgreenbg,
  colframe=darkgreen,
  boxrule=0.4pt,
  arc=1.5mm,
  left=1.5mm,right=1.5mm,top=1mm,bottom=1mm
]
The white and black kings.
\end{tcolorbox}

\textbf{Model output}
\begin{tcolorbox}[
  colback=lightredbg,
  colframe=darkred,
  boxrule=0.4pt,
  arc=1.5mm,
  left=1.5mm,right=1.5mm,top=1mm,bottom=1mm,
  breakable
]
The model interprets the gift as a decorative object made with African materials such as ebony and ivory. It hypothesizes that the missing elements were symbols of slavery (chains and shackles), which may have been removed to avoid political awkwardness given Jefferson's role and the ideals of liberty.

\texttt{<Answer>chains and shackles</Answer>}
\end{tcolorbox}

\textbf{Predicted answer:} \textcolor{darkred}{chains and shackles}

\end{tcolorbox}

\begin{tcolorbox}[
  colback=white,
  colframe=bordergray,
  boxrule=0.5pt,
  arc=2mm,
  left=2mm,right=2mm,top=1mm,bottom=1mm,
  title=\textbf{Incorrect concept anchoring}
]

\textbf{Question}
\begin{tcolorbox}[
  colback=lightbluebg,
  colframe=darkblue,
  boxrule=0.4pt,
  arc=1.5mm,
  left=1.5mm,right=1.5mm,top=1mm,bottom=1mm
]
Michael Chabon compared a river glistening with metal, whose waves crash between the embankments, to an X. The internet project ``X,'' conceived by Boris Berezovsky, was intended to report on violations of civil liberties in Russia. Name X.
\end{tcolorbox}

\textbf{Gold answer}
\begin{tcolorbox}[colback=lightgreenbg,colframe=darkgreen,boxrule=0.4pt,arc=1.5mm]
Bell.
\end{tcolorbox}

\textbf{Model output}
\begin{tcolorbox}[
  colback=lightredbg,
  colframe=darkred,
  boxrule=0.4pt,
  arc=1.5mm,
  breakable
]
The model interprets the metaphor as referring to a zipper (``molniya'' in Russian), and connects this to a supposed internet project named ``Molniya'' associated with Boris Berezovsky.

\texttt{<Answer>Molniya</Answer>}
\end{tcolorbox}

\textbf{Predicted answer:} \textcolor{darkred}{Molniya}

\end{tcolorbox}

\begin{tcolorbox}[
  colback=white,
  colframe=bordergray,
  boxrule=0.5pt,
  arc=2mm,
  left=2mm,right=2mm,top=1mm,bottom=1mm,
  title=\textbf{Incorrect concept anchoring}
]

\textbf{Question}
\begin{tcolorbox}[
  colback=lightbluebg,
  colframe=darkblue,
  boxrule=0.4pt,
  arc=1.5mm
]
In his descriptions, the author of the novel mentions the radiance of a rainbow, often adorning the head, as well as the crown of a weeping willow. Name this author.
\end{tcolorbox}

\textbf{Gold answer}
\begin{tcolorbox}[colback=lightgreenbg,colframe=darkgreen,boxrule=0.4pt,arc=1.5mm]
Herman Melville.
\end{tcolorbox}

\textbf{Model output}
\begin{tcolorbox}[
  colback=lightredbg,
  colframe=darkred,
  boxrule=0.4pt,
  arc=1.5mm,
  breakable
]
The model interprets the clues as referring to the butterfly \textit{Apatura iris} and associates this imagery with Vladimir Nabokov, known for his interest in butterflies and descriptive prose.

\texttt{<Answer>Vladimir Nabokov</Answer>}
\end{tcolorbox}

\textbf{Predicted answer:} \textcolor{darkred}{Vladimir Nabokov}

\end{tcolorbox}

\begin{tcolorbox}[
  colback=white,
  colframe=bordergray,
  boxrule=0.5pt,
  arc=2mm,
  left=2mm,right=2mm,top=1mm,bottom=1mm,
  title=\textbf{Incorrect concept anchoring}
]

\textbf{Question}
\begin{tcolorbox}[
  colback=lightbluebg,
  colframe=darkblue,
  boxrule=0.4pt,
  arc=1.5mm
]
In a humorous translation of the first part of the film ``The Lord of the Rings,'' Gandalf says to Frodo: ``I love your Hobbiton — the nature here is beautiful.'' In Gandalf's line, we replaced two words with one. Write the replaced words.
\end{tcolorbox}

\textbf{Gold answer}
\begin{tcolorbox}[colback=lightgreenbg,colframe=darkgreen,boxrule=0.4pt,arc=1.5mm]
New Zealand.
\end{tcolorbox}

\textbf{Model output}
\begin{tcolorbox}[
  colback=lightredbg,
  colframe=darkred,
  boxrule=0.4pt,
  arc=1.5mm,
  breakable
]
The model refers to a humorous Russian translation and interprets ``Hobbiton'' as a placeholder for the phrase ``Tikhiy Omut'' (``still waters''), concluding that the replaced words are ``Still waters.''

\texttt{<Answer>Still waters</Answer>}
\end{tcolorbox}

\textbf{Predicted answer:} \textcolor{darkred}{Still waters}

\end{tcolorbox}

\begin{tcolorbox}[
  colback=white,
  colframe=bordergray,
  boxrule=0.5pt,
  arc=2mm,
  left=2mm,right=2mm,top=1mm,bottom=1mm,
  title=\textbf{Wrong hypothesis}
]

\textbf{Question}
\begin{tcolorbox}[
  colback=lightbluebg,
  colframe=darkblue,
  boxrule=0.4pt,
  arc=1.5mm
]
There are currently four such countries on the world map, although there used to be many more. Among them are Egypt, France, and Spain. Name the fourth country.
\end{tcolorbox}

\textbf{Gold answer}
\begin{tcolorbox}[colback=lightgreenbg,colframe=darkgreen,boxrule=0.4pt,arc=1.5mm]
Panama.
\end{tcolorbox}

\textbf{Model output}
\begin{tcolorbox}[
  colback=lightredbg,
  colframe=darkred,
  boxrule=0.4pt,
  arc=1.5mm,
  breakable
]
The model identifies a shared property among Egypt, France, and Spain as countries bordering the Mediterranean Sea and having an additional coastline. It constructs a geographical explanation involving countries with dual coastlines and concludes that Morocco fits this pattern.

\texttt{<Answer>Morocco</Answer>}
\end{tcolorbox}

\textbf{Predicted answer:} \textcolor{darkred}{Morocco}

\end{tcolorbox}

\begin{tcolorbox}[
  colback=white,
  colframe=bordergray,
  boxrule=0.5pt,
  arc=2mm,
  left=2mm,right=2mm,top=1mm,bottom=1mm,
  title=\textbf{Wrong reference}
]

\textbf{Question}
\begin{tcolorbox}[
  colback=lightbluebg,
  colframe=darkblue,
  boxrule=0.4pt,
  arc=1.5mm
]
An alien from Zelazny's novel gnaws off pieces of a diamond brick due to the peculiarities of its physiology. What did a character from an 1892 story involuntarily use for similar purposes?
\end{tcolorbox}

\textbf{Gold answer}
\begin{tcolorbox}[colback=lightgreenbg,colframe=darkgreen,boxrule=0.4pt,arc=1.5mm]
Blue carbuncle.
\end{tcolorbox}

\textbf{Model output}
\begin{tcolorbox}[
  colback=lightredbg,
  colframe=darkred,
  boxrule=0.4pt,
  arc=1.5mm,
  breakable
]
The model interprets the question through a biological analogy involving continuously growing teeth, drawing a parallel between the alien and animals such as hares. It references a 19th-century story featuring a frightened hare and concludes that the character used his own teeth involuntarily.

\texttt{<Answer>His own teeth</Answer>}
\end{tcolorbox}

\textbf{Predicted answer:} \textcolor{darkred}{His own teeth}

\end{tcolorbox}

\begin{tcolorbox}[
  colback=white,
  colframe=bordergray,
  boxrule=0.5pt,
  arc=2mm,
  left=2mm,right=2mm,top=1mm,bottom=1mm,
  title=\textbf{Clue misinterpretation}
]

\textbf{Question}
\begin{tcolorbox}[
  colback=lightbluebg,
  colframe=darkblue,
  boxrule=0.4pt,
  arc=1.5mm
]
Sculptor Henry Moore equated women with nature, so \textbf{DEFINITIONS} in his works could symbolize ravines or depressions eroded by the forces of wind and water. What two words did we replace with the word ``DEFINITIONS''?
\end{tcolorbox}

\textbf{Gold answer}
\begin{tcolorbox}[colback=lightgreenbg,colframe=darkgreen,boxrule=0.4pt,arc=1.5mm]
Curvy lines.
\end{tcolorbox}

\textbf{Model output}
\begin{tcolorbox}[
  colback=lightredbg,
  colframe=darkred,
  boxrule=0.4pt,
  arc=1.5mm,
  breakable
]
The model focuses on Henry Moore’s use of negative space and interprets the description as referring to hollow spaces or holes within sculptures, drawing an analogy to natural formations such as caves or valleys.

\texttt{<Answer>hollow spaces</Answer>}
\end{tcolorbox}

\textbf{Predicted answer:} \textcolor{darkred}{hollow spaces}

\end{tcolorbox}
\begin{table*}[h]
\centering
\small
\scalebox{0.82}{
\begin{tabular}{p{1.8cm} p{1.2cm} p{4.8cm} p{2.4cm} p{4.8cm}}
\toprule
\textbf{Difficulty} & \textbf{ID} & \textbf{Question} & \textbf{Answer} & \textbf{Explanation} \\
\midrule

\rowcolor{green!8}
Very Easy &
\texttt{f12464c4c7} &
The Turkic poet Saif Sarai in the poem ``Sukhail and Guldursun'' metaphorically describes a girl circling around her illustrious beloved. Whom, according to the translator, did Sarai precede by about one and a half centuries? &
[Nicolaus] Copernicus. &
Sarai (a 14th-century poet) compared the beauty Guldursun to the Earth revolving around the Sun. Copernicus is considered the discoverer of heliocentrism. \\[0.5em]

\rowcolor{yellow!8}
Easy &
\texttt{b338b9556c} &
In the game ``Virtua Tennis 4'' instead of a tennis ball, a bomb is used: whoever ends up with the X on their side loses. Who created ``X'' in 1986? &
The band ``Europe.'' &
Countdown. ``Final Countdown'' is a famous song and album by the band "Europe. \\[0.5em]

\rowcolor{blue!6}
Medium &
\texttt{83a712ed75} &
In a wartime novel, the hero looks at the sky, where contrails of airplanes and bursts of anti-aircraft shells form an unpleasant, in his opinion, picture. What word did we replace with ``picture''? &
Melody. &
Contrails are the lines of a musical staff, and the bursts are the symbols of notes. \\[0.5em]

\rowcolor{red!8}
Hard &
\texttt{d124832645} &
The name of this American mountain, on the crack of which a climber is crawling, matches the title of a cult 1983 film. Write this name. &
``Scarface'' &
The rock resembles a human profile, and the crack is like a scar on this face. \\[0.5em]

\rowcolor{red!12}
Very Hard &
\texttt{f65f89f5bd} &
In the description of a silent scene, Mikhail Belousov mentions a sound produced by THEM. ``THEIR'' sound was recognized as the best 36 years ago. Name THEM. &
Jaws. &
A silent scene is usually the result of extreme astonishment and is accompanied by jaws dropping. The film ``Jaws'' received numerous awards for best sound and music. \\[0.5em]

\bottomrule
\end{tabular}
}
\caption{Examples from our benchmark at each difficulty level with the gold answer and the explanation.}
\label{tab:benchmark-examples}
\end{table*}

\rowcolors{3}{gray!15}{white}

\begin{figure}[h]
    \centering
    \begin{minipage}{0.32\textwidth}
        \centering
        \includegraphics[width=\linewidth]{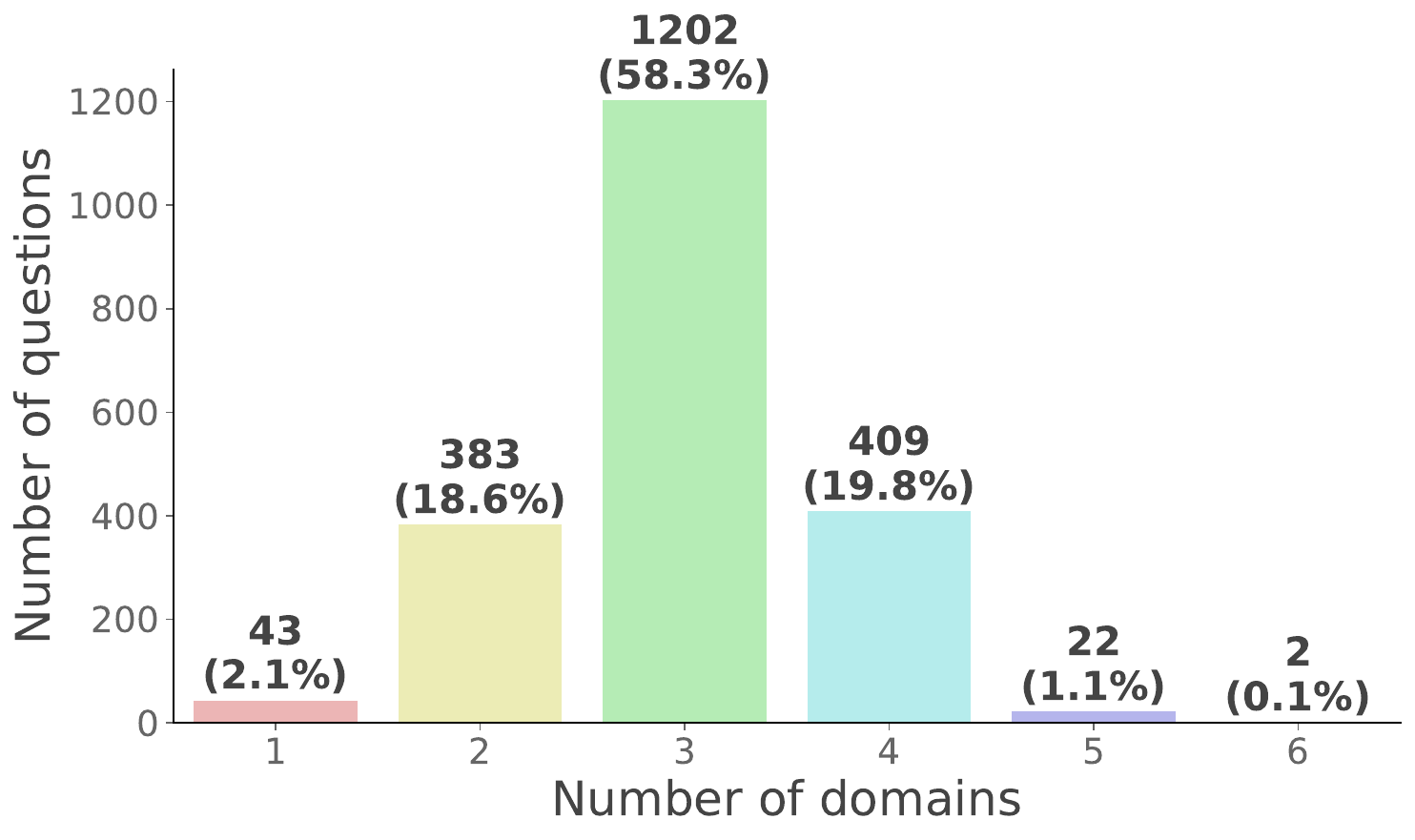}
        \subcaption{Knowledge domains.\label{fig:num-domains}}
    \end{minipage}
    \hfill
    \begin{minipage}{0.32\textwidth}
        \centering
        \includegraphics[width=\linewidth]{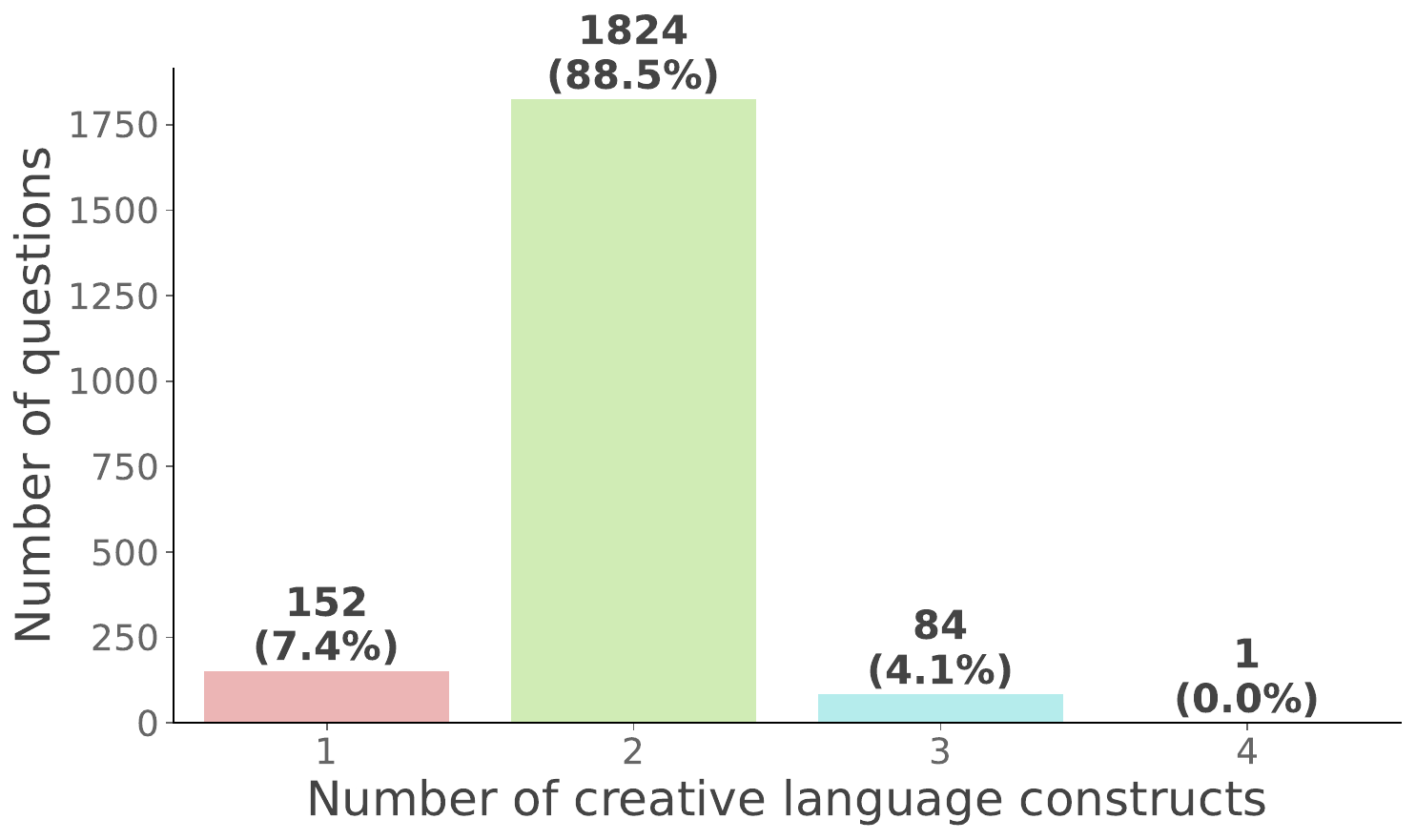}
        \subcaption{Creativity concepts.\label{fig:num-concepts}}
    \end{minipage}
    \hfill
    \begin{minipage}{0.32\textwidth}
        \centering
        \includegraphics[width=\linewidth]{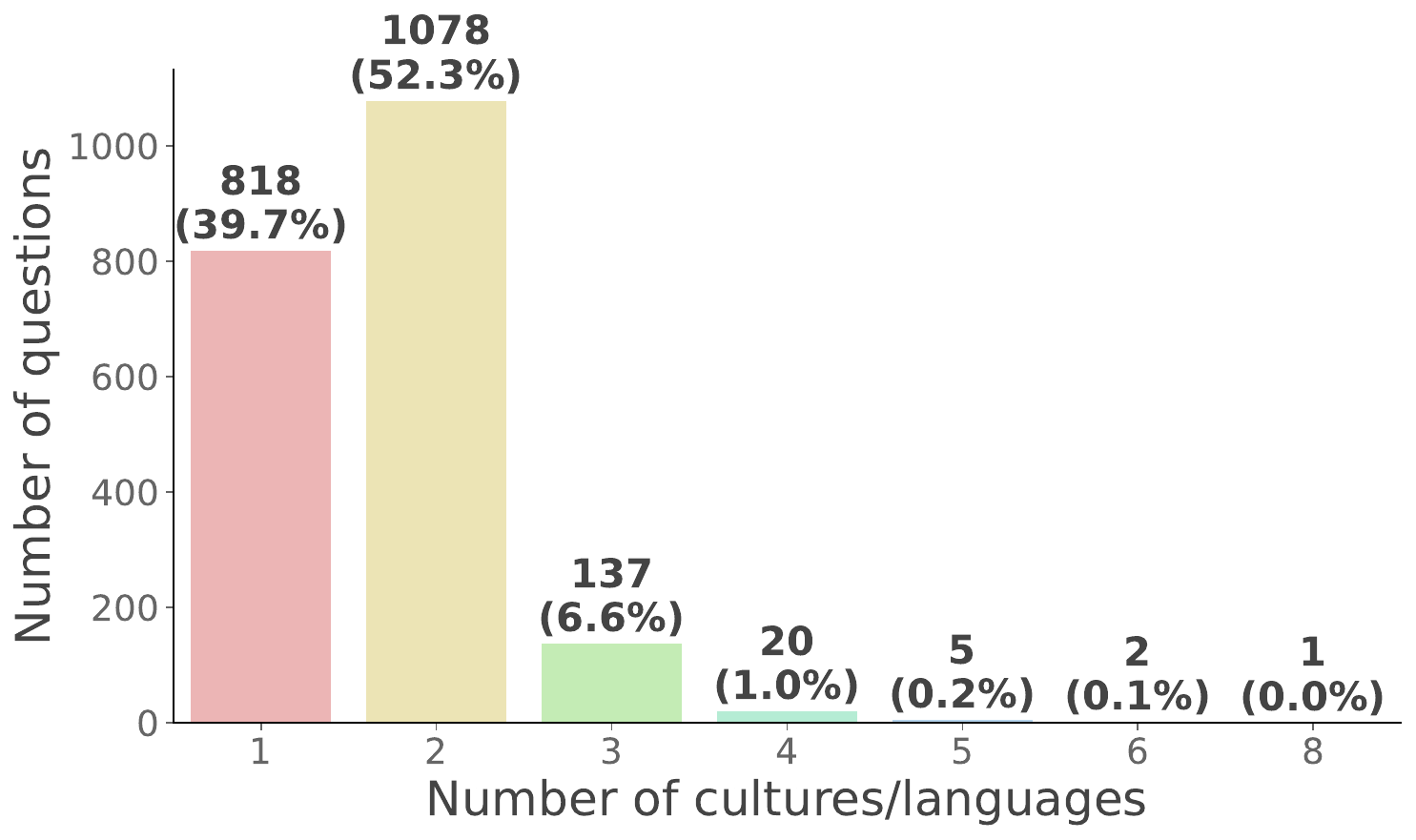}
        \subcaption{Cultures/demographics.\label{fig:num-langs}}
    \end{minipage}
    \caption{Distribution of number of knowledge domains, creative language constructs and cultures/languages per question.}
    \label{fig:num-domain-cre-langs}
\end{figure}

\begin{table*}[t]
\centering
\small
\scalebox{0.82}{
\begin{tabular}{p{4cm} p{1cm} p{1cm} p{1cm} p{1.5cm} p{1.5cm} p{1.5cm} p{1.5cm}}
& \multicolumn{7}{c}{\textbf{Features}} \\
\toprule
\textbf{Benchmarks} & \textbf{Conv.} & \textbf{Div.} & \textbf{Lat.} & \textbf{Analogy} & \textbf{Com.Sense} & \textbf{Real-World KG} & \textbf{Domains} \\
\midrule
\textbf{BRAINTEASER} \citep{jiang2023brainteaser} & \CheckmarkBold & \CheckmarkBold & \CheckmarkBold & \textcolor{red}{\XSolidBrush} & \CheckmarkBold & \textcolor{red}{\XSolidBrush} & Various \\
\textbf{RiddleSense} \citep{lin2021riddlesense} & \CheckmarkBold & \textcolor{red}{\XSolidBrush} & \textcolor{red}{\XSolidBrush} & \textcolor{red}{\XSolidBrush} & \CheckmarkBold & \CheckmarkBold & Various \\
\textbf{Briangle} \cite{han2025creativity} & \textcolor{red}{\XSolidBrush} & \CheckmarkBold & \CheckmarkBold & \textcolor{red}{\XSolidBrush} & \textcolor{red}{\XSolidBrush} & \textcolor{red}{\XSolidBrush} & Math/Logic \\
\textbf{SPLAT} \citep{chen2024weak} & \textcolor{red}{\XSolidBrush} & \CheckmarkBold & \CheckmarkBold & \textcolor{red}{\XSolidBrush} & \CheckmarkBold & \textcolor{red}{\XSolidBrush} & Daily-Life \\
\textbf{LatEval} \citep{huang2024lateval} & \textcolor{red}{\XSolidBrush} & \CheckmarkBold & \CheckmarkBold & \textcolor{red}{\XSolidBrush} & \CheckmarkBold & \textcolor{red}{\XSolidBrush} & Daily-Life \\
\textbf{Columbus} \citep{kraaijveld2025columbus} & \CheckmarkBold & \CheckmarkBold & \CheckmarkBold & \textcolor{red}{\XSolidBrush} & \textcolor{red}{\XSolidBrush} & \textcolor{red}{\XSolidBrush} & Logic \\
\textbf{Connections} \citep{todd2024missed} & \textcolor{red}{\XSolidBrush} & \CheckmarkBold & \CheckmarkBold & \textcolor{red}{\XSolidBrush} & \CheckmarkBold & \textcolor{red}{\XSolidBrush} & Various \\
\textbf{DAT} \citep{bellemare2024divergent} & \textcolor{red}{\XSolidBrush} & \CheckmarkBold & \textcolor{red}{\XSolidBrush} & \textcolor{red}{\XSolidBrush} & \textcolor{red}{\XSolidBrush} & \textcolor{red}{\XSolidBrush} & Various \\
\textbf{McGyver} \citep{tian2024macgyver} & \CheckmarkBold & \CheckmarkBold & \textcolor{red}{\XSolidBrush} & \textcolor{red}{\XSolidBrush} & \CheckmarkBold & \CheckmarkBold & Daily-Life \\
\textbf{OnlyConnect} \citep{alavi2023large} & \CheckmarkBold & \textcolor{red}{\XSolidBrush} & \textcolor{red}{\XSolidBrush} & \textcolor{red}{\XSolidBrush} & \CheckmarkBold & \CheckmarkBold & Various \\
\textbf{CREATE} \citep{wadhwa2026createtestingllmsassociative} & \CheckmarkBold & \CheckmarkBold & \textcolor{red}{\XSolidBrush} & \textcolor{red}{\XSolidBrush} & \textcolor{red}{\XSolidBrush} & \CheckmarkBold & Various \\
\citep{lewis2024using} & \CheckmarkBold & \textcolor{red}{\XSolidBrush} & \textcolor{red}{\XSolidBrush} & \CheckmarkBold & \textcolor{red}{\XSolidBrush} & \textcolor{red}{\XSolidBrush} & Logic \\
\textbf{NEOCODER}
\citep{lu2025benchmarking} & \CheckmarkBold & \CheckmarkBold & \textcolor{red}{\XSolidBrush} & \textcolor{red}{\XSolidBrush} & \textcolor{red}{\XSolidBrush} & \textcolor{red}{\XSolidBrush} & Coding \\
\textbf{CreativeMath}
\citep{ye2025assessing} & \CheckmarkBold & \CheckmarkBold & \textcolor{red}{\XSolidBrush} & \textcolor{red}{\XSolidBrush} & \textcolor{red}{\XSolidBrush} & \textcolor{red}{\XSolidBrush} & Math \\
\textbf{OMEGA}
\citep{sun2025omega} & \CheckmarkBold & \textcolor{red}{\XSolidBrush} & \textcolor{red}{\XSolidBrush} & \textcolor{red}{\XSolidBrush} & \textcolor{red}{\XSolidBrush} & \textcolor{red}{\XSolidBrush} & Math \\
\textbf{\ourbench{}} (Ours) & \CheckmarkBold & \CheckmarkBold & \CheckmarkBold & \CheckmarkBold & \CheckmarkBold & \CheckmarkBold & Various \\
\bottomrule
\end{tabular}
}
\caption{Comparison of different creative problem-solving benchmarks. Our benchmark notably requires employing several creative thinking strategies and involves various domains. (Conv.=Convergent, Div.=Divergent, Lat.=Lateral, Com.Sense=Commonsense reasoning)}
\label{tab:cps-benchmarks}
\end{table*}

\begin{table*}[t]
\centering
\small
\scalebox{0.82}{
\begin{tabular}{p{1.8cm} p{1.2cm} p{4.8cm} p{4.8cm} p{2.4cm}}
\toprule
\textbf{Pipeline Step} & \textbf{ID} & \textbf{Question (Russian)} & \textbf{Question (English)} & \textbf{Answer} \\
\midrule

\multicolumn{5}{l}{\textit{\textbf{Step 1: Filtering unanswerable questions} (removed 295 samples)}} \\[0.3em]

\rowcolor{red!8}
Requires physical materials &
\texttt{7e149044a7} &
\foreignlanguage{russian}{Статья в ``Нью-Йорк Таймс'', посвященная идее возврата к золотому стандарту в целях укрепления доллара, сопровождалась изображением, фрагмент которого мы вам раздали. Воспроизведите то, что мы закрыли черным прямоугольником.} &
An article in ``The New York Times'' about the idea of returning to the gold standard to strengthen the dollar was accompanied by an image, a fragment of which we distributed to you. Reproduce what we covered with a black rectangle. &
In God We Trust \\[0.5em]

\midrule
\multicolumn{5}{l}{\textit{\textbf{Step 2: Filtering Russian-specific questions} (removed 799 samples)}} \\[0.3em]

\rowcolor{red!8}
Russian language/ culture specific &
\texttt{fbce8ea84b} &
\foreignlanguage{russian}{В произведении Евгения Лукина ОН в ответ на вопрос героя призывает отрицать эпидемию. Не спрашиваем, какие четыре слова мы заменили словами ``отрицать эпидемию''. Назовите ЕГО.} &
In the work of Yevgeny Lukin, IT, in response to the hero's question, urges to deny the epidemic. We do not ask which four words we replaced with ``deny the epidemic.'' Name IT. &
\foreignlanguage{russian}{Ворон} / Raven \newline \scriptsize{\textit{(Pun on Poe's Raven: ``\foreignlanguage{russian}{Не верь в мор!}'' --- ``Don't believe in plague!'')}} \\[0.5em]

\midrule
\multicolumn{5}{l}{\textit{\textbf{Step 3: Translation} (all remaining 2,695 samples translated with \texttt{GPT-4o})}} \\[0.3em]

\rowcolor{blue!6}
Translated &
\texttt{c97e6fd16d} &
\foreignlanguage{russian}{По предположению американских ученых, складки, которые образуются на мокрых пальцах, выполняют ту же функцию, что и ИКС. После смерти отца ИКСОМ был провозглашен Ричард. Назовите его фамилию.} &
According to American scientists, the wrinkles that form on wet fingers serve the same function as X. After the death of his father, Richard was proclaimed X. Name his surname. &
\foreignlanguage{russian}{Кромвель} / Cromwell \newline \scriptsize{\textit{(X = Lord Protector; wrinkles protect grip.)}} \\[0.5em]

\midrule
\multicolumn{5}{l}{\textit{\textbf{Step 4: Human validation} (removed 282 samples)}} \\[0.3em]

\rowcolor{red!8}
Still Russian-specific &
\texttt{506e7ac07e} &
\foreignlanguage{russian}{Старейший американский клуб, основанный в 1903 году, называется ``Белые медведи Кони-Айленда''. Сезон у них длится с ноября по апрель. Каким словом мы называем участников подобных клубов?} &
The oldest American club, founded in 1903, is called the ``Coney Island Polar Bears.'' Their season lasts from November to April. What word do we use to refer to members of such clubs? &
\foreignlanguage{russian}{Моржи} / Walruses \newline \scriptsize{\textit{(Russian-specific term for winter swimmers.)}} \\[0.3em]

\rowcolor{red!8}
Requires external materials &
\texttt{17cad876d2} &
\foreignlanguage{russian}{Перед вами фрагмент альтернативного постера к известному фильму. Назовите этот фильм.} &
In front of you is a fragment of an alternative poster for a famous film. Name this film. &
\foreignlanguage{russian}{``Кофе и сигареты''} / ``Coffee and Cigarettes'' \\[0.3em]

\rowcolor{red!8}
Multiple questions &
\texttt{e089848e4f} &
\foreignlanguage{russian}{Блиц. 1. На рекламной этикетке художника Баранова есть надпись ``С кружкой пивца и работается!'' и изображение человека. Назовите этого человека. [...]} &
Blitz. 1. On the advertising label by artist Baranov, there is an inscription ``With a mug of beer, work is dear!'' and an image of a person. Name this person. [...] &
1. \foreignlanguage{russian}{Ленин} / Lenin \newline 2. \foreignlanguage{russian}{Ленин} / Lenin \newline 3. \foreignlanguage{russian}{Ленин} / Lenin \\[0.3em]

\rowcolor{red!8}
Outdated info &
\texttt{c95beab3bb} &
\foreignlanguage{russian}{Михаил Прохоров, отвечая на неудобный вопрос, сказал, что ЕВА у него появилась в 17 лет. Назовите имя и фамилию нынешней американской ЕВЫ.} &
Mikhail Prokhorov, answering an awkward question, said that EVA appeared in his life at the age of 17. Name the first and last name of the current American EVA. &
\foreignlanguage{russian}{Мишель Обама} / Michelle Obama \newline \scriptsize{\textit{(Answer is outdated.)}} \\[0.5em]

\midrule
\multicolumn{5}{l}{\textit{\textbf{Step 5: Creative vs.\ factual annotation} (352 factual, 2,061 creative)}} \\[0.3em]

\rowcolor{green!8}
Factual &
\texttt{8947f691ac} &
\foreignlanguage{russian}{Вольф Шнайдер пишет, что один кристалл ЕГО в 1970-х годах лишил работы 45 тысяч швейцарцев. Назовите ЕГО.} &
Wolf Schneider writes that one crystal of IT in the 1970s deprived 45,000 Swiss people of their jobs. Name IT. &
\foreignlanguage{russian}{Кварц} / Quartz \newline \scriptsize{\textit{(Quartz watches replaced Swiss mechanical watches.)}} \\[0.3em]

\rowcolor{yellow!12}
Creative &
\texttt{83a712ed75} &
\foreignlanguage{russian}{В романе о военном времени герой смотрит на небо, где инверсионные следы самолетов и разрывы зенитных снарядов образуют неприятную, по его мнению, картину. Какое слово мы заменили на ``картину''?} &
In a wartime novel, the hero looks at the sky, where contrails of airplanes and bursts of anti-aircraft shells form an unpleasant, in his opinion, picture. What word did we replace with ``picture''? &
\foreignlanguage{russian}{Мелодию} / Melody \newline \scriptsize{\textit{(Contrails are staff lines; bursts are notes.)}} \\

\bottomrule
\end{tabular}
}
\caption{Examples of questions affected at each stage of the data processing pipeline. For each step, we show a representative question in Russian (original) and English (translated), along with its answer and the reason for filtering or annotation.}
\label{tab:pipeline-examples}
\end{table*}

\begin{table*}[ht]
\centering
\begin{tcolorbox}[title=External Material Filtering Prompt, breakable, colback=white, colframe=black!70, fonttitle=\bfseries\small, width=\textwidth]
\small\ttfamily
You are a strict annotator. Your task is to decide whether the given puzzle EXPLICITLY instructs the participant to check external materials to answer the question.\par\medskip
Answer "yes" ONLY if the puzzle requires explicit external resource lookup, such as:\par
\hspace{1em}-- A handout \foreignlanguage{russian}{(``раздаточный материал'')} or statements like \foreignlanguage{russian}{``на розданной вам фотографии'', ``смотрите раздатку'', ``не озвучивать текст раздатки'', ``в раздатке'',} or similar \par
\hspace{1em}-- References to a hidden or closed element such as: \foreignlanguage{russian}{``мы закрыли символ'', ``мы закрыли букву'', ``мы закрыли часть текста'', ``прочерк'',}, when the missing content is NOT provided within the puzzle text.\par
\hspace{1em}-- Instructions requiring external lookup such as \foreignlanguage{russian}{``зайдите по ссылке и посмотрите курс'', ``вставьте число из источника''} etc.\par\medskip
Answer "no" if:\par
\hspace{1em}-- The puzzle does not EXPLICITLY instruct the participant to check external materials.\par\medskip
Output ONLY 'yes' or 'no' in lowercase, with no additional text.\par\medskip
Puzzle:\par
\{question\}\par\medskip
Additional notes:\par
\{comment\}\par\medskip
Answer:
\end{tcolorbox}
\caption{Zero-shot prompt used to filter questions requiring physical external materials (See \S\ref{sec:data_filtering}) for details.}
\label{tab:prompt-ext-mat}
\end{table*}

\begin{table*}[ht]
\centering
\begin{tcolorbox}[title=Translation Prompt, breakable, colback=white, colframe=black!70, fonttitle=\bfseries\small, width=\textwidth]
\small\ttfamily
You are a professional Russian to English translator. Your task is to translate Russian puzzles into English with absolute fidelity.\par\medskip
Translate EXACTLY, preserving:\par
\hspace{1em}-- all logical clues\par
\hspace{1em}-- all named entities\par
\hspace{1em}-- sentence order and structure\par
\hspace{1em}-- rhetorical devices\par
\hspace{1em}-- ambiguity\par
\hspace{1em}-- references and style\par
\hspace{1em}-- the puzzle's original difficulty\par\medskip
Do NOT:\par
\hspace{1em}-- paraphrase\par
\hspace{1em}-- simplify or summarize\par
\hspace{1em}-- interpret hidden meanings\par
\hspace{1em}-- add explanations\par
\hspace{1em}-- rephrase stylistically\par\medskip
If a segment is unclear, translate it literally.\par\medskip
You are given the puzzle question, answer, comment and notes in Russian enclosed in <Question>, <Answer>, <Comment>, and <Notes> tags.\par
Translate each section into English, preserving the structure and meaning, and output the translated puzzle in the same format with <Question>, <Answer>, <Comment>, and <Notes> tags.\par\medskip
Output ONLY the English translation, nothing else.\par\medskip
Russian Puzzle:\par
<Question>\{question\}</Question>\par
<Answer>\{answer\}</Answer>\par
<Comment>\{comment\}</Comment>\par
<Notes>\{notes\}</Notes>\par\medskip
English Translation:
\end{tcolorbox}
\caption{Zero-shot prompt used for translating puzzles from Russian to English (\S\ref{sec:translation}).}
\label{tab:prompt-translation}
\end{table*}

\begin{table*}[ht]
\centering
\begin{tcolorbox}[title=Russian-Specific Filtering Prompt (Instruction), breakable, colback=white, colframe=black!70, fonttitle=\bfseries\small, width=\textwidth]
\small\ttfamily
You are a strict annotator. You are given a puzzle in Russian, along with its answer, comment, and notes and translation of the puzzle into English. Your task is to decide whether the puzzle can not be solved in English because solving the puzzle REQUIRES knowledge specific to the Russian LANGUAGE.\par\medskip
Answer 'yes' ONLY if the solution depends on Russian-specific linguistic features such as:\par
\hspace{1em}-- idioms, sayings, or set expressions that break in translation\par
\hspace{1em}-- wordplay, puns, or jokes that work ONLY in Russian\par
\hspace{1em}-- Russian-specific homonyms or near-homonyms\par
\hspace{1em}-- phonetic or rhyming clues that exist ONLY in Russian\par
\hspace{1em}-- letter-based tricks involving the Russian alphabet or spelling\par
\hspace{1em}-- translating the puzzle into another language would change or destroy the meaning, making the puzzle unsolvable or fundamentally different.\par\medskip
Answer 'no' if:\par
\hspace{1em}-- the puzzle depends on general knowledge, culture, history, geography, literature, or context, even if these are Russian.\par
\hspace{1em}-- the puzzle uses Russian names/entities but no Russian linguistic tricks\par
\hspace{1em}-- the translated puzzle still is solvable without losing the key idea.\par\medskip
Given the puzzle question, answer, comment, notes in Russian and the English translation of the puzzle, output your reasoning in English within <Reasoning>...</Reasoning> tags and your answer within <Answer>...</Answer> tags.
\end{tcolorbox}

\begin{tcolorbox}[title=Russian-Specific Filtering Prompt (Few-Shot Example Format), colback=gray!5, colframe=black!50, fonttitle=\bfseries\small, width=\textwidth]
\small\ttfamily
Russian puzzle:\par
Question: \{question\}\par
Answer: \{answer\}\par
Comment: \{comment\}\par
Notes: \{notes\}\par\medskip
English translation:\par
Question: \{question\_en\}\par
Answer: \{answer\_en\}\par
Comment: \{comment\_en\}\par
Notes: \{notes\_en\}\par\medskip
<Reasoning>\{reasoning\}</Reasoning>\par
<Answer>\{shot\_answer\}</Answer>
\end{tcolorbox}
\caption{Few-shot prompt used to filter Russian-specific questions (See \S\ref{sec:data_filtering} for details).}
\label{tab:prompt-ru-spec}
\end{table*}

\begin{table*}[ht]
\centering
\begin{tcolorbox}[title=Creative vs.\ Factual Reasoning Annotation Prompt (Instruction), breakable, colback=white, colframe=black!70, fonttitle=\bfseries\small, width=\textwidth]
\small\ttfamily
You are an expert in annotating type of reasoning involved in solving creative thinking puzzles. Given a creative thinking puzzle question, its answer, comments about the puzzle and other acceptable answers (if any), do TWO tasks:\par\medskip
1) Decide whether solving the puzzle mainly requires SIMPLE FACTUAL reasoning or CREATIVE ASSOCIATION of distant pieces of knowledge/facts.\par
\hspace{1em}-- Answer 'factual' when the puzzle is answered by simply retrieving specific facts, dates, names, or straightforward commonsense reasoning, with no need to make a non-obvious, creative leaps between facts.\par
\hspace{1em}-- Answer 'creative' when the solver in addition to retrieving the facts, must also make a non-obvious, creative connection, see a hidden twist, reinterpret words, or combine clues in an indirect way.\par\medskip
2) If your answer to the previous task is 'CREATIVE', then choose ALL relevant creativity concepts involved in solving the puzzle from the following list (use EXACT spelling, lowercase):\par
poem, metaphor, idiom, proverb, joke, pun, simile, sarcasm, hyperbole, neologism, analogy, abstraction, lateral thinking, divergent thinking, commonsense reasoning, compositionality.\par
\hspace{1em}-- Choose ALL that apply (multi-label).\par
\hspace{1em}-- If none of the specific labels fit, suggest a list of new labels that are relevant for the puzzle.\par\medskip
Carefully read the given puzzle and output:\par
\hspace{1em}-- One line with the reasoning type: <Answer>factual</Answer> OR <Answer>creative</Answer>.\par
\hspace{1em}-- If you chose 'creative', on separate lines output ONE OR MORE <concept> tags, each containing either exactly one label from the allowed list or a new label if none fits.\par
\hspace{1em}-- If you chose 'factual', do NOT output any <concept> tags.
\end{tcolorbox}

\begin{tcolorbox}[title=Creative vs.\ Factual Reasoning Annotation Prompt (Few-Shot Example Format), colback=gray!5, colframe=black!50, fonttitle=\bfseries\small, width=\textwidth]
\small\ttfamily
Example \{index\}:\par
Puzzle:\par
\{question\}\par
Answer: \{answer\}\par
Comment: \{comment\}\par
Acceptable Answers: \{notes\}\par
Annotation:\par
\{shot\_answer\}
\end{tcolorbox}
\caption{Few-shot prompt used for creative vs.\ factual reasoning annotation ( See \S\ref{sec:creative_vs_factual} for details).}
\label{tab:prompt-reasoning-type}
\end{table*}

\begin{table*}[ht]
\centering
\begin{tcolorbox}[title=Domain Annotation Prompt (Instruction), breakable, colback=white, colframe=black!70, fonttitle=\bfseries\small, width=\textwidth]
\small\ttfamily
You are an expert in annotating creative thinking puzzles. Given a creative thinking puzzle, its answer, comments about the puzzle and other acceptable answers (if any), identify a list of domains that are involved in solving the puzzle. A domain is a specific area of knowledge, expertise, or human activity such as physics, literature, sports, music, etc. Wrap each domain on a separate line in <domain>...</domain> tags.
\end{tcolorbox}

\begin{tcolorbox}[title=Domain Annotation Prompt (Few-Shot Example Format), colback=gray!5, colframe=black!50, fonttitle=\bfseries\small, width=\textwidth]
\small\ttfamily
Example \{index\}:\par
Puzzle:\par
\{question\}\par
Answer: \{answer\}\par
Comment: \{comment\}\par
Acceptable Answers: \{notes\}\par
Domains:\par
\{domains\}
\end{tcolorbox}
\caption{Few-shot prompt used for knowledge domain annotation (\S\ref{sec:benchmark}).}
\label{tab:prompt-domain}
\end{table*}

\begin{table*}[ht]
\centering
\begin{tcolorbox}[title=Knowledge Annotation Prompt (Instruction), breakable, colback=white, colframe=black!70, fonttitle=\bfseries\small, width=\textwidth]
\small\ttfamily
You are an expert in annotating creative thinking puzzles. Given a creative thinking puzzle, identify a list of knowledge/facts that are explicitly required to answer the puzzle, and write them in the form of independent questions. Don't solve the problem. Don't include the answer itself. Wrap each question on a separate line in <knowledge>...</knowledge> tags.
\end{tcolorbox}

\begin{tcolorbox}[title=Knowledge Annotation Prompt (Few-Shot Example Format), colback=gray!5, colframe=black!50, fonttitle=\bfseries\small, width=\textwidth]
\small\ttfamily
Example \{index\}:\par
Puzzle:\par
\{question\}\par
Answer: \{answer\}\par
Comment: \{comment\}\par
Acceptable Answers: \{notes\}\par
Knowledge:\par
\{knowledge\}
\end{tcolorbox}
\caption{Few-shot prompt used for knowledge decomposition annotation (See \S\ref{sec:diff-complexity}).}
\label{tab:prompt-knowledge}
\end{table*}

\begin{table*}[ht]
\centering
\begin{tcolorbox}[title=Culture/Demographics Annotation Prompt (Instruction), breakable, colback=white, colframe=black!70, fonttitle=\bfseries\small, width=\textwidth]
\small\ttfamily
You are an expert in annotating creative thinking puzzles. Given a creative thinking puzzle, its answer, comments about the puzzle and other acceptable answers (if any), identify the languages/cultures that are involved in solving the puzzle. Output only your final answer and separate them with commas.
\end{tcolorbox}

\begin{tcolorbox}[title=Culture/Demographics Annotation Prompt (Few-Shot Example Format), colback=gray!5, colframe=black!50, fonttitle=\bfseries\small, width=\textwidth]
\small\ttfamily
Example \{index\}:\par
Puzzle:\par
\{question\}\par
Answer: \{answer\}\par
Comment: \{comment\}\par
Acceptable Answers: \{notes\}\par
Cultures/Languages: \{cultures\}
\end{tcolorbox}
\caption{Few-shot prompt used for culture and demographics annotation (\S\ref{sec:cultures}).}
\label{tab:prompt-culture-lang}
\end{table*}

\begin{table*}[ht]
\centering
\begin{tcolorbox}[title=Chain-of-Thought Prompt (English), breakable, colback=white, colframe=black!70, fonttitle=\bfseries\small, width=\textwidth]
\small\ttfamily
You are an expert at solving creative thinking puzzles. You are given a puzzle question that requires creative reasoning over real-world knowledge. Note that the question might contain several placeholder words (often in all capital case such as X, Y, THIS, FIRST, SECOND, ALPHA, BETA, HE, SHE, IT, HIM, HIS, HER, THEY, THEM etc.) that substitute for specific entities, objects, or concepts. These placeholders are crucial for solving the puzzle, and their meaning can only be inferred through careful reasoning about the question. Additionally, some placeholders might be gendered (e.g., HE/HIM/HIS vs SHE/HER), but do not assume that the gendered pronouns necessarily refer to human characters; they could refer to any entities, and their gender might be different. Given a puzzle question, think step by step and write your reasoning within <Reasoning>...</Reasoning> tags. Then provide only the final answer within <Answer>...</Answer> tags.\par
Puzzle:\par
\{question\}
\end{tcolorbox}

\begin{tcolorbox}[title=Chain-of-Thought Prompt (Russian), colback=white, colframe=black!70, fonttitle=\bfseries\small, width=\textwidth]
\small\ttfamily
You are an expert at solving creative thinking puzzles. You are given a puzzle question in Russian that requires creative reasoning over real-world knowledge. Note that the question might contain several placeholder words (often in all capital case such as \foreignlanguage{russian}{ИКС, ИГРЕК, ЭТО, ПЕРВЫЙ, ВТОРОЙ, АЛЬФА, БЕТА, ОН, ОНА, ЕГО, ЕЕ, ОНИ} etc.) that substitute for specific entities, objects, or concepts. These placeholders are crucial for solving the puzzle, and their meaning can only be inferred through careful reasoning about the question. Given a puzzle question, think step by step and write your reasoning in Russian within <Reasoning>...</Reasoning> tags. Then provide only the final answer in Russian (unless specified otherwise) within <Answer>...</Answer> tags.\par
Puzzle:\par
\{question\}
\end{tcolorbox}

\begin{tcolorbox}[title=Thinking Model Prompt (English), colback=white, colframe=black!70, fonttitle=\bfseries\small, width=\textwidth]
\small\ttfamily
You are an expert at solving creative thinking puzzles. You are given a puzzle question that requires creative reasoning over real-world knowledge. Note that the question might contain several placeholder words (often in all capital case such as X, Y, THIS, FIRST, SECOND, ALPHA, BETA, HE, SHE, IT, HIM, HIS, HER, THEY, THEM etc.) that substitute for specific entities, objects, or concepts. These placeholders are crucial for solving the puzzle, and their meaning can only be inferred through careful reasoning about the question. Additionally, some placeholders might be gendered (e.g., HE/HIM/HIS vs SHE/HER), but do not assume that the gendered pronouns necessarily refer to human characters; they could refer to any entities, and their gender might be different. Please provide your final answer within <Answer>...</Answer> tags.\par
Puzzle:\par
\{question\}
\end{tcolorbox}

\begin{tcolorbox}[title=Thinking Model Prompt (Russian), colback=gray!5, colframe=black!70, fonttitle=\bfseries\small, width=\textwidth]
\small\ttfamily
You are an expert at solving creative thinking puzzles. You are given a puzzle question in Russian that requires creative reasoning over real-world knowledge. Note that the question might contain several placeholder words (often in all capital case such as \foreignlanguage{russian}{ИКС, ИГРЕК, ЭТО, ПЕРВЫЙ, ВТОРОЙ, АЛЬФА, БЕТА, ОН, ОНА, ЕГО, ЕЕ, ОНИ} etc.) that substitute for specific entities, objects, or concepts. These placeholders are crucial for solving the puzzle, and their meaning can only be inferred through careful reasoning about the question. Please provide your final answer in Russian (unless specified otherwise) within <Answer>...</Answer> tags.\par
Puzzle:\par
\{question\}
\end{tcolorbox}
\caption{Evaluation prompts for non-thinking (CoT) and thinking (reasoning) models in English and Russian (\S\ref{sec:evaluation}).}
\label{tab:prompt-evaluation}
\end{table*}

\begin{table*}[ht]
\centering
\begin{tcolorbox}[title=LLM-as-a-Judge Prompt, breakable, colback=white, colframe=black!70, fonttitle=\bfseries\small, width=\textwidth]
\small\ttfamily
You are an expert in judging model responses. Given a question, a reference answer, a model answer, and additional context such as comments and other acceptable answers, decide whether the model's answer is correct based on the reference answer and context. Ignore minor typos, articles, capitalization, and formatting. If the factual core is the same, answer yes. Also ignore the number of words constraint in the question if the model answer is semantically correct but does not meet the word count requirement. Answer only: Yes or No.\par\medskip
Question:\par
\{question\}\par
Reference Answer:\par
\{answer\}\par
Additional Context:\par
Comments: \{comment\}\par
Acceptable Answers: \{notes\}\par
Model Answer:\par
\{prediction\}
\end{tcolorbox}
\caption{Zero-shot prompt used for LLM-as-a-judge evaluation (\S\ref{sec:metrics}).}
\label{tab:prompt-llm-judge}
\end{table*}

\begin{table*}[ht]
\centering
\begin{tcolorbox}[title=Data Contamination Perturbation Prompt, breakable, colback=white, colframe=black!70, fonttitle=\bfseries\small, width=\textwidth]
\small\ttfamily
Your task is to create a three-choice quiz by replacing the words in the provided ``Question'' with their contextually relevant synonyms.
The meaning and sentence structure of the three options must exactly match every detail in the Question. You must not include the provided Question as an option. You must make sure that:\par
(1) You generate distinct options based on the provided Question;\par
(2) The only difference between options is word-level perturbations;\par
(3) There is not any extra explanation;\par
(4) You comply with every specific symbol and letter detail in the given Question; and\par
Given a Question, output three alternative question options and enclose each option with <option>...</option> tags.\par
Question:\{question\}\par
Options:
\end{tcolorbox}
\caption{Zero-shot prompt used for data contamination pertubation annotation (\S\ref{sec:contamination}).}
\label{tab:prompt-contamination-perturb}
\end{table*}

\begin{table*}[ht]
\centering
\begin{tcolorbox}[title=Data Contamination Quiz Prompt, breakable, colback=white, colframe=black!70, fonttitle=\bfseries\small, width=\textwidth]
\small\ttfamily
You are provided with a four-choice quiz. Your task is to correctly select the option that exactly corresponds to an instance from the What?Where?When (\foreignlanguage{russian}{Что? Где? Когда?}, also abbreviated as \foreignlanguage{russian}{ЧГК} or ChGK) game dataset.\par
(1) You must ensure that you only generate a single option letter as your answer.\par
(2) If you do not know the dataset or the correct answer, you must select option D) None of the provided options.\par
Hint: While all the following options seem similar in terms of meaning, there is only one option that reflects the exact wording with respect to the original instance.\par
A) Question: \{option\_a\}\par
B) Question: \{option\_b\}\par
C) Question: \{option\_c\}\par
D) None of the provided options.\par
Answer:
\end{tcolorbox}
\caption{Zero-shot prompt used for data contamination quiz (\S\ref{sec:contamination}).}
\label{tab:prompt-contamination-quiz}
\end{table*}

\begin{figure}[h]
\includegraphics[width=\linewidth, trim={0cm 0cm 0cm 0cm},clip]{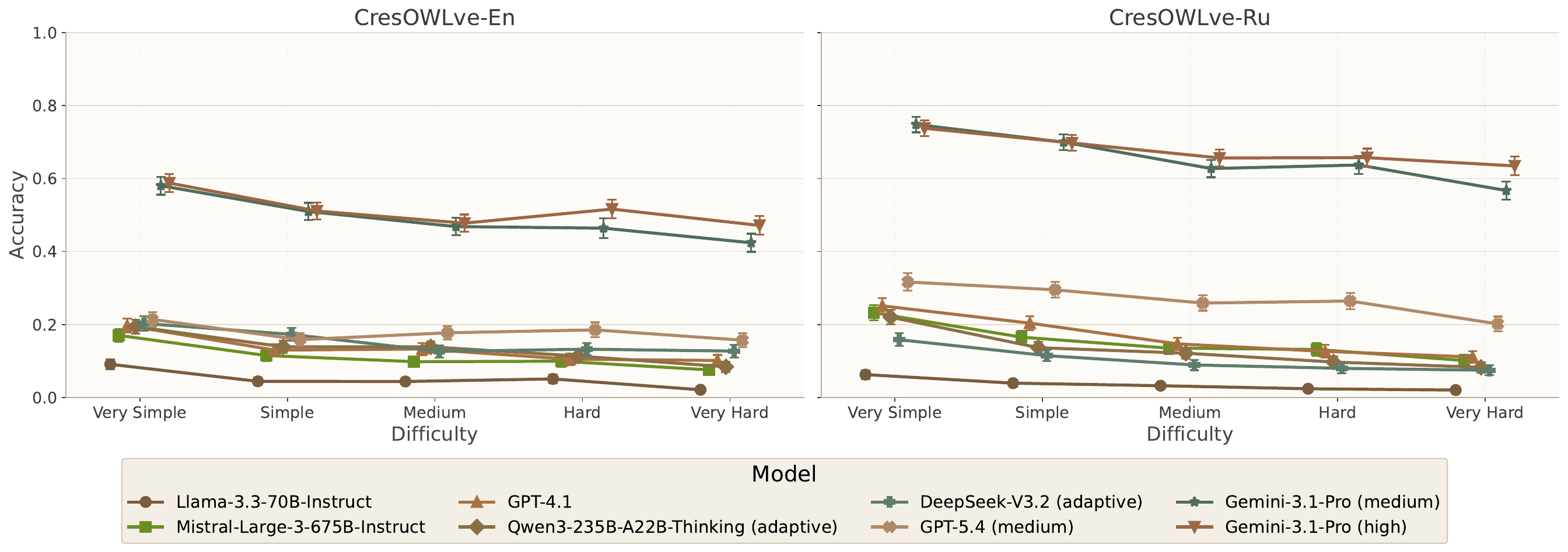}
\caption{Exact Match Performance by difficulty.}
\label{fig:diff-results-em}
\end{figure}

\begin{figure}[h]
\includegraphics[width=\linewidth, trim={0cm 0cm 0cm 0cm},clip]{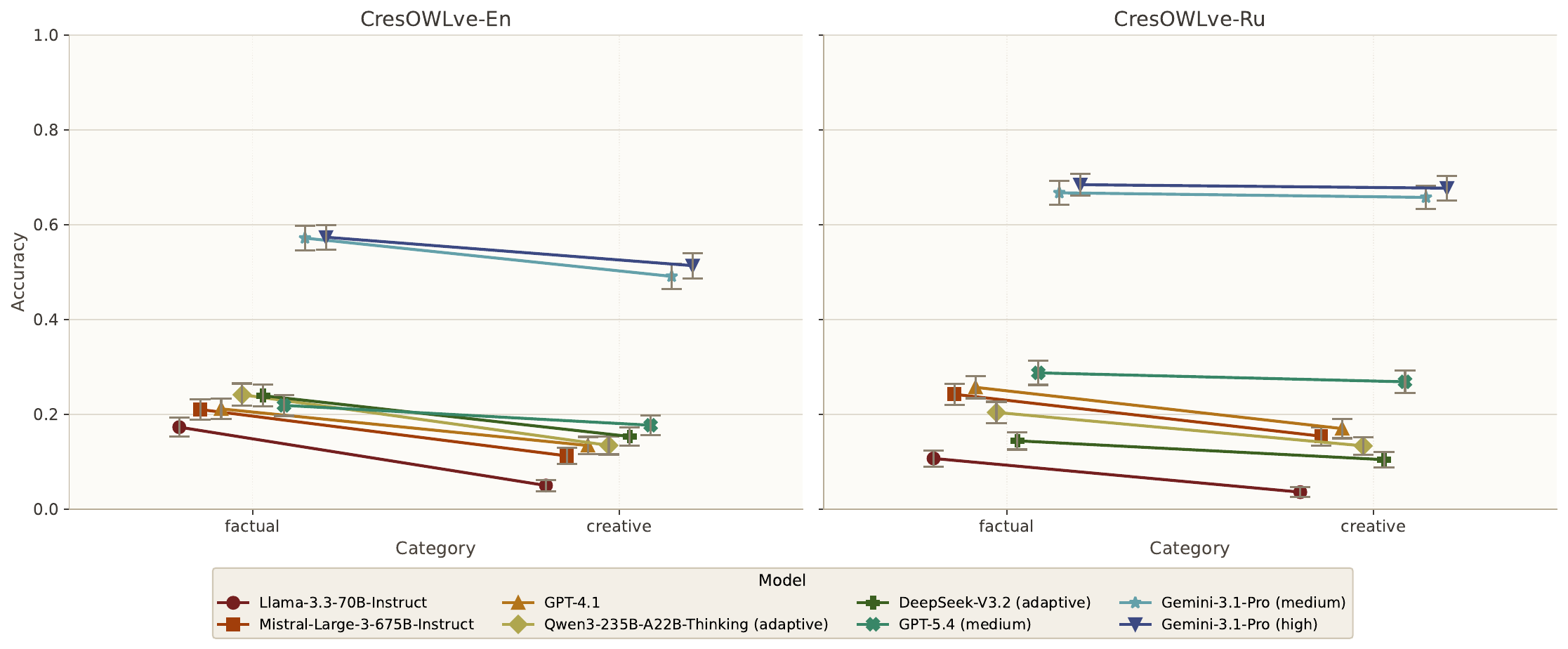}
\caption{Exact Match Performance by reasoning category.}
\label{fig:cre-fac-results-em}
\end{figure}

\begin{figure}[h]
    \centering
    \begin{minipage}{0.45\textwidth}
        \centering
        \includegraphics[width=\linewidth]{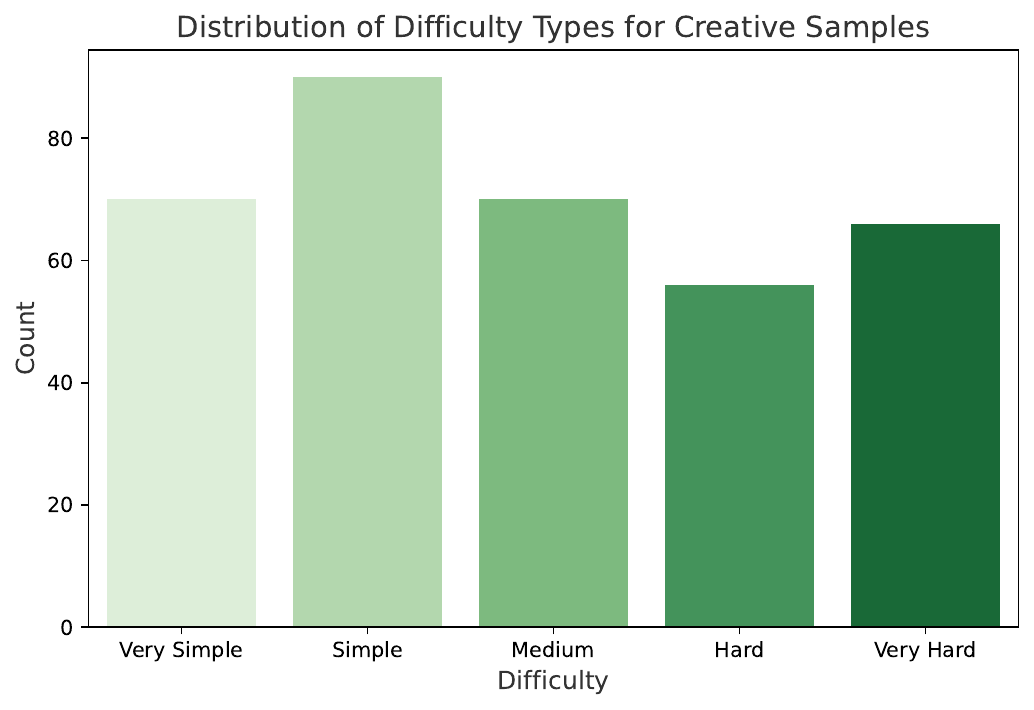}
    \end{minipage}
    \hfill
    \begin{minipage}{0.45\textwidth}
        \centering
        \includegraphics[width=\linewidth]{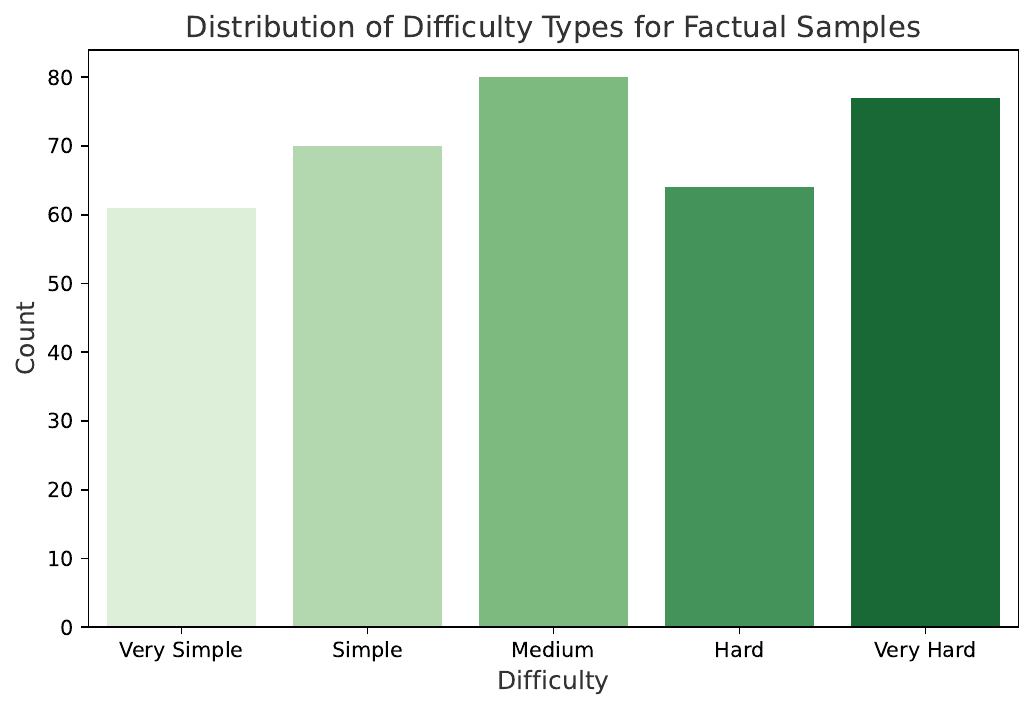}
    \end{minipage}
    \caption{Distribution of difficulty levels for creative and factual questions.}
    \label{fig:creative-factual-diff-dist}
\end{figure}

\begin{figure}[t]
\includegraphics[width=\linewidth, trim={0cm 0cm 0cm 0cm},clip]{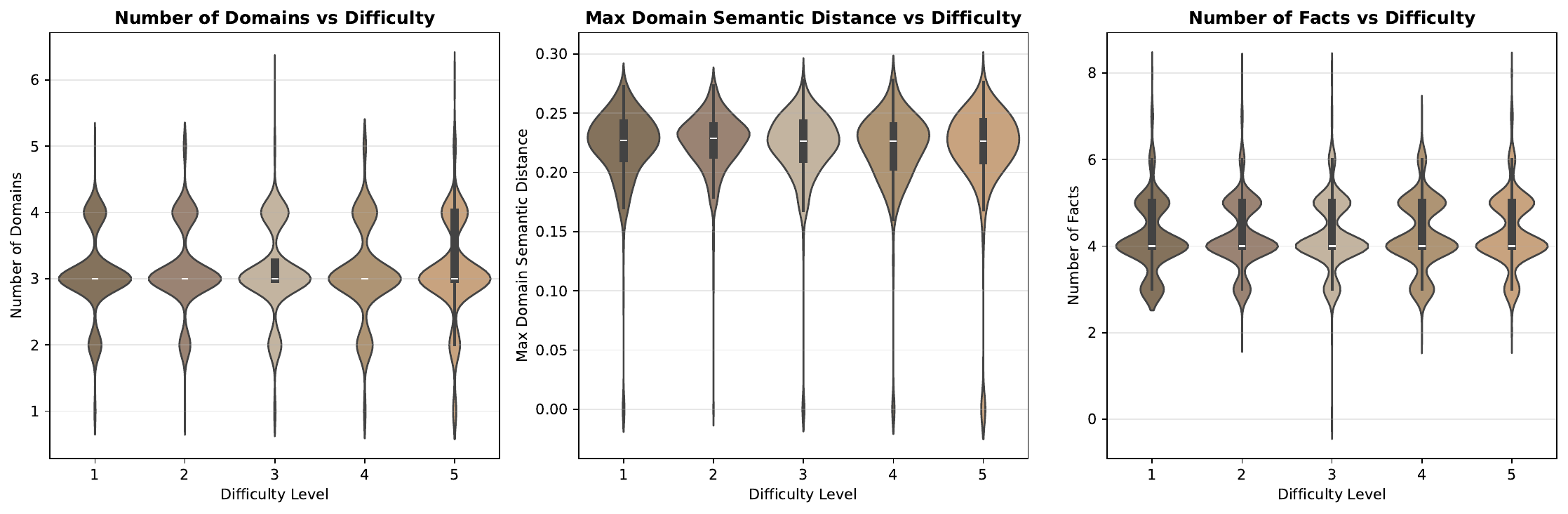}
\caption{Correlations between question difficulty and complexity features.}
\label{fig:diff-corrs}
\end{figure}

\begin{figure}[t]
\includegraphics[width=\linewidth, trim={0cm 0cm 0cm 0cm},clip]{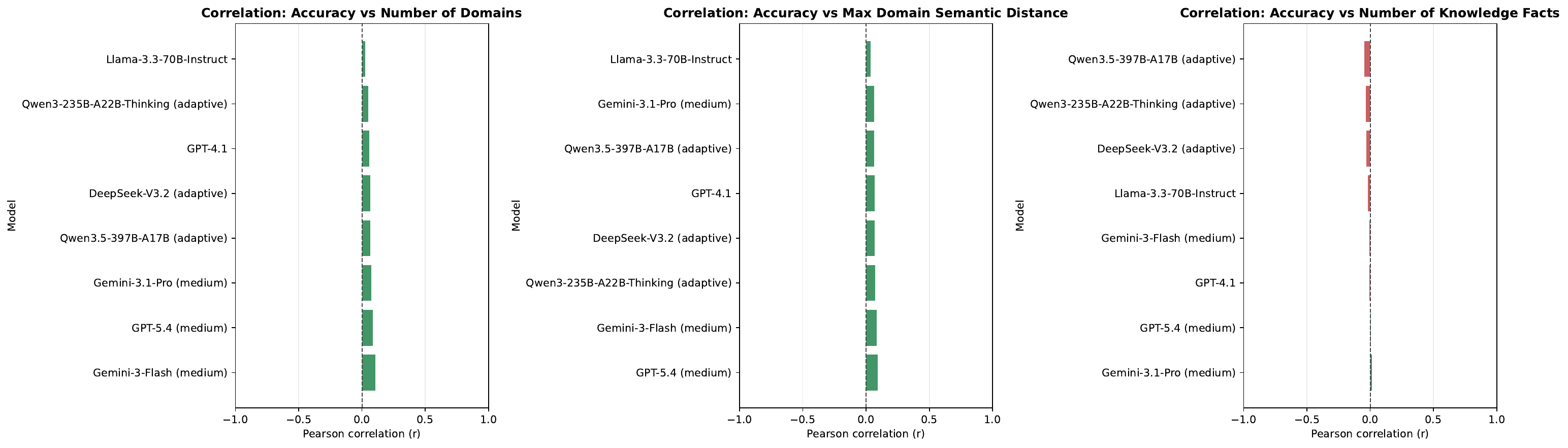}
\caption{Correlations between model performance and complexity features.}
\label{fig:perf-corrs}
\end{figure}

\end{document}